# Spatial Heuristic Strategy based Diffusion Reinforcement Learning for Online 3D Bin Packing

Tong Li, Qingyang Xu+

School of Airspace Science and Engineering, Shandong University, Weihai 264209, China, and also with the Shandong Key Laboratory of Intelligent Electronic Packaging Testing and Application, Shandong University, Weihai 264209.

+ Corresponding author.

Email:202317537@mail.sdu.edu.cn and qingyangxu @sdu.edu.cn

**Abstract:** Traditional heuristic methods have accumulated mature experience in core objectives, such as space utilization and layout compactness of 3D bin packing, relying on artificially designed rule-based logic. However, they exhibit obvious limitations in generalization ability and adaptability to dynamic scenarios. Diffusion reinforcement learning, which integrates the spatial search capability of reinforcement learning and the data distribution modeling capability of diffusion learning, has achieved certain progress in action distribution modeling and decision compliance constraints. Nevertheless, there remain issues including insufficient perception of the spatial topological structure in 3D bin packing and limited layout optimization capability. To address these problems, we propose a diffusion reinforcement learning method integrated with spatial heuristic strategies, which combines classical bin packing heuristic rules with a diffusion reinforcement learning framework. Specifically, it compresses the action space through heuristic action pre-screening, introduces a graph neural network to model the spatial structure, and designs a composite reward function incorporating flatness to suppress spatial fragmentation, thereby realizing the organic integration of prior knowledge and adaptive optimization. Experimental results demonstrate that the proposed algorithm effectively enhances the performance and stability of online 3D bin packing.

## 1 Introduction

In practical scenarios such as logistics, warehousing, and manufacturing, how to efficiently load items within confined spaces stands as a core issue for improving resource utilization and system operational efficiency. As an important research direction in the field of spatial optimization, the Bin Packing Problem (BPP)[1] has long attracted extensive attention from the domains of operations research and artificial intelligence. Three-dimensional bin packing represents a typical NP-hard problem, and most mature current solution approaches are centered on heuristic strategies. Such methods rely on the prior knowledge of researchers, who manually construct guiding rules and decision criteria to achieve efficient exploration of feasible packing spaces. Consequently, approximate optimal solutions satisfying business constraints are commonly adopted in real engineering applications[2–5].

Heuristic methods exhibit certain intuitiveness and effectiveness in solving the three-dimensional bin packing problem. However, restricted by manually designed rules, such methods cannot theoretically guarantee global optimal solutions, and their performance is difficult to maintain consistently across different problem scenarios. In contrast, Deep Reinforcement Learning (DRL) has shown significant advantages in addressing such combinatorial optimization problems [6]. Zhang et al. [7] proposed a brain-inspired experience reinforcement model by imitating human empirical reasoning, enabling adaptation to complex and variable environments. Que et al.[8] developed a DRL model based on the Transformer architecture [9] to solve the offline three-dimensional bin packing problem with variable height, yet this method struggles to adjust strategies in real time to cope with unexpected situations.

Zhang et al. [10] presented Attend2Pack, a deep reinforcement learning-based method incorporating an attention mechanism, which overcomes the limitations of traditional algorithms in handling high dimensionality and complex constraints. Cheng et al. [11] proposed an end-to-end deep reinforcement learning approach fusing Transformer and CNN, alleviating the excessive action space issue in online multi-bin three-dimensional packing. It improves space utilization while maintaining uniform height distribution of bins. Zhao et al. [12] addressed the deficiencies in observation modeling and the challenge of sparse rewards in three-dimensional bin packing by proposing the DMRL-BPP dynamic multimodal deep reinforcement learning framework. It employs twin encoders to capture multimodal information, designs a novel reward function to optimize training feedback, and combines the SAC algorithm with heuristic placement strategies to reduce the action space. This framework outperforms heuristic algorithms and mainstream DRL methods in terms of gap ratio and learning stability across different bin sizes, significantly enhancing packing space utilization efficiency. Pan et al. [13] tackled the trade-off between average performance and worst-case performance in online three-dimensional bin packing by proposing an adjustable robust reinforcement learning framework. It generates worst-case scenarios through permutation-based attackers and balances the two types of performance via a weighted objective function, improving worst-case space utilization and stability without obvious degradation in normal performance. Nevertheless, online three-dimensional packing methods based on reinforcement learning still suffer from issues such as blind exploration of the action space, insufficient utilization of layout prior knowledge, poor training stability, and low sample efficiency.

Diffusion models possess powerful data generation capabilities. By progressively removing noise to generate reasonable packing action distributions, diffusion models can not only capture the complex multimodal distributions in the action space, but also dynamically adjust the action generation strategy according to the current environmental state. Introducing diffusion models into the actor-critic framework to construct a diffusion-model-based AC reinforcement learning method enables modeling of high-dimensional packing action distributions, thereby enhancing the representation and exploration capabilities of complex layouts [14]. Although diffusion models have achieved certain progress in action distribution modeling and decision-making compliance constraints, they still suffer from insufficient perception of the spatial topological structure in 3D bin packing and limited layout optimization ability under complex scenarios. Despite the obvious limitations of heuristic methods in generalization and dynamic scene adaptability, they have accumulated mature optimization experience on core objectives such as space utilization and layout compactness through manually designed rules and logic. Yang et al. [15] systematically discussed how to embed three categories of heuristic methods—physical heuristics, packing heuristics, and unpacking heuristics—into the DRL framework, constructing a heuristic-guided deep reinforcement learning model that significantly improves the practicality and generalization of policies for online 3D-BPP. Xiong et al. [16] introduced a candidate graph generated based on heuristics to achieve efficient search in the large discrete action space. Aiming at the problems of poor performance, difficult training, and insufficient adaptability in existing methods for online 3D bin packing, Zhang et al. [16] proposed a heuristic-guided enhanced data deep reinforcement learning method, which improves packing space utilization in both discrete and continuous scenarios with more stable performance and higher training efficiency. Although diffusion reinforcement learning has great potential for adaptive decision-making, it is difficult to directly inherit the prior knowledge of heuristic rules, resulting in insufficient efficiency in exploring efficient layouts during the training process. Heuristic methods can utilize rich prior knowledge but suffer from insufficient generalization. Therefore, this paper proposes a diffusion reinforcement learning method integrated with spatial heuristic strategies, which combines classical bin packing heuristic rules with a data-driven reinforcement learning framework. It extracts four types of heuristic rules, such as maximum subspace division and minimum remaining space, to generate candidate action solutions. Meanwhile, it introduces a graph neural network to mine the topological associations between goods and space, and designs a composite reward function incorporating flatness penalty to optimize the regularity of packing layouts, thereby realizing the organic integration of prior knowledge and adaptive optimization.

## 2 Related works

The solution to the three-dimensional bin packing problem based on heuristic algorithms can effectively utilize prior knowledge and is often adopted in practical engineering applications. However, it has the problem of poor generalization ability. To seek a better optimization strategy, researchers have begun to attempt to solve the three-dimensional bin packing problem using DRL-based methods [8,17]. Zhu et al. [36] proposed a data-driven tree search algorithm, which uses tree search to explore the complex solution space and combines a convolutional neural network trained on historical data to guide pruning, thereby accelerating the search process. Zhang et al. [7] proposed a brain-inspired experience reinforcement model by imitating human empirical reasoning, enabling it to adapt to complex and variable environments. Que et al. [8] developed a DRL model based on the Transformer[9] architecture to solve the offline three-dimensional bin packing with variable height, but this method is difficult to adjust strategies in real time to cope with unexpected situations. In contrast, the online bin packing problem focused on in this paper is more of practical research significance.

In recent years, research on DRL in the field of online 3D-BPP has been continuously in-depth. Jia et al. [18] combined DRL with Monte Carlo Tree Search (MCTS), used future item information to assist decision-making, and effectively alleviated the convergence problem under high-dimensional discrete action space through the improved Actor-Critic algorithm [19] and heuristic rule model. On this basis, Zhao et al. [20] constructed a constrained DRL framework and introduced a feasibility mask prediction module to improve learning efficiency and policy quality. Their subsequent work [21] proposed a packing configuration tree as a structured representation of the state and action space, simplified DRL policy learning, and performed excellently in the continuous solution space. In addition, some studies have focused on combining heuristic rules with reinforcement learning to improve the practicality and generalization ability of policies. Yang et al. [15] systematically discussed how to embed three categories of heuristic methods—physical, packing, and unpacking heuristics—into the DRL framework, constructing a heuristic-guided deep reinforcement learning model that significantly improves the policy practicality and generalization ability of online 3D-BPP. Xiong et al.[16] introduced a candidate graph generated based on heuristics to achieve efficient search in the large discrete action space and successfully migrated the model to practical robotic systems. Song et al.[22] incorporated an unpacking mechanism into the DRL framework to learn the synergistic effect of packing and unpacking behaviors, allowing the agent to correct improper operations, thereby improving the utilization rate of packing policies. Tsang et al. [23] addressed the problems of single-bin focus, difficulty in dual-bin adaptation, and complex action space in online three-dimensional bin packing, proposing a double deep Q-network optimizer. It constructs states through the maximum cuboid algorithm, designs a composite reward function, decomposes packing actions, and presents two bin replacement strategies. Experiments have shown that it improves space utilization in both single-bin and dual-bin scenarios, with more significant advantages under low lookahead numbers, providing an efficient solution for automated packing in robotic warehouses.

The hierarchical Advantage Actor-Critic (A2C) reinforcement learning framework proposed by Zhang et al. [24] is based on this concept and is specifically applied to online airline luggage packing scenarios. This framework first decomposes the overall packing problem into multi-level subtasks, then independently optimizes each level using the A2C algorithm, thereby achieving step-by-step optimization from top to bottom. More importantly, the feedback information obtained from the bottom-level optimization can in turn adjust the upper-level strategy, ultimately achieving the goal of global optimization. Another important study comes from Liu et al. [25], who proposed a model called "chunk-based dual-task learning". This model cleverly uses deep learning technology to simultaneously learn to perform two closely related tasks: determining the optimal orientation and placement position of each item in the bin. By exploring and utilizing the commonalities between these two tasks, the model can significantly improve generalization ability and overall performance in different packing scenarios. The Attend2Pack method proposed by Zhang et al. [10] takes deep reinforcement learning as the core and subtly incorporates an attention mechanism, bringing an innovative solution to the three-dimensional bin packing problem. This method effectively breaks through the limitations of traditional algorithms in handling high dimensionality and complex constraints, providing a new deep learning paradigm based on reinforcement learning for packing optimization problems. Cheng et al. [11] proposed an end-to-end deep reinforcement learning method fusing Transformer and CNN, which effectively solves the problem of excessive action space in online multi-bin three-dimensional packing, improving space utilization while ensuring uniform height distribution of bins. To address the deficiencies in observation modeling and the challenge of sparse rewards in three-dimensional bin packing, Zhao et al. [12] proposed the DMRL-BPP dynamic multimodal deep reinforcement learning framework. It uses twin encoders to capture multimodal information, designs a new reward function to optimize training feedback, and combines the SAC algorithm with heuristic placement strategies to reduce the action space. Experiments have shown that it significantly improving packing space utilization efficiency. Aiming at the problems of poor performance, difficult training, and insufficient adaptability of existing methods for online three-dimensional bin packing, Zhang et al. [26] proposed a heuristic-guided enhanced data deep reinforcement learning method, achieving better packing space utilization with more stable performance and higher training efficiency. Hu et al. [27] addressed the "Transport-and-Pack (TAP)" problem in reality, where items are occluded and need to be transported in sequence before packing, proposing a deep reinforcement learning-based neural network model TAP-Net. This model cleverly uses a priority graph to represent the occlusion relationship between items and combines an attention mechanism to end-to-end learn the optimal item transportation sequence and packing strategy. It has shown excellent performance and good generalization ability in both 2D and 3D scenarios, providing a new idea for solving packing problems under complex constraints. Li et al. [28] proposed the RCQL model, which uses the Transformer model to handle the offline three-dimensional bin packing problem. The packing process is divided into three consecutive sub-actions, and the policy network generates policies for these three sub-actions respectively. On this basis, Jiang et al. [29] further used action representation learning to solve the large action space of the position selection sub-action, significantly improving space utilization. Pan et al. [13] tackled the pain point that it is difficult to balance average performance and worst-case performance in online three-dimensional bin packing, proposing an adjustable robust reinforcement learning framework. It generates worst-case scenarios through permutation-based attackers and balances the two types of performance with a weighted objective function. Experiments have shown that this framework can greatly improve the space utilization and stability of worst-case scenarios on various datasets without significantly sacrificing normal performance.

Yin et al. [30] addressed the contradiction between efficiency and sequence adjustment in the semi-online three-dimensional bin packing problem, as well as the landing challenges of reinforcement learning solutions, proposing a new semi-online packing framework. Through the adjustable order setting of observable multiple items, a Transformer network adapting to a variable number of items, and a bottom-up hybrid reward function, it achieves better packing space utilization than baseline methods while ensuring the feasibility of robotic arm movement. Yang et al. [31] proposed the PUSNet packing-unpacking collaborative framework, introducing an unpacking error correction mechanism and large language model guidance, which greatly improves the performance of online three-dimensional packing and achieves the current optimal packing effect. Puche et al. [20] addressed the problems of low decision-making freedom and insufficient sample efficiency in online three-dimensional bin packing, for the first time adapting AlphaGo-like model-based reinforcement learning methods to the three-dimensional bin packing task, proposing a reinforcement learning framework with a buffer zone. It expands the decision space through the buffer zone and improves sample efficiency through symmetric data augmentation, ultimately achieving better packing space utilization with lower training hardware requirements. Wu et al. [32] addressed the bottleneck of the reinforcement learning action space in robotic palletizing scenarios, proposing an iterative action mask learning method. Through the iteratively optimized learning-based action mask, it accurately prunes the action space, which not only accelerates the training process of reinforcement learning but also improves the final packing space utilization, with performance significantly superior to traditional heuristic mask methods. Zhang et al. [33] addressed the challenge of missing physical properties and reliance on action mask rules in robotic palletizing scenarios, proposing a physically-aware online action mask framework. It incorporates the physical properties of bins into decision-making and gets rid of reliance on manual rules through the online-learned action mask, ultimately achieving performance surpassing existing methods and being deployable to real palletizing robots. Yuan et al. [34] addressed the bottleneck of reinforcement learning training variance in industrial decision-making tasks, proposing a dual-critic "demand-constraint" decoupling method. By decoupling environmental dynamics to reduce gradient variance, it achieves better space utilization and faster training convergence in online three-dimensional bin packing tasks, which can also be generalized to other industrial tasks. Ahn et al. [35] addressed the real landing pain points of mixed palletizing, proposing a practical system fusing the PMP-RL palletizing model and CMPNet motion planning. It balances the stability and space utilization of packing, as well as the real-time performance and collision-free movement of the robotic arm, and can be directly deployed to real warehouse scenarios.

## 3 Spatial Heuristic Strategy Based Markov Modeling of 3D Bin Packing

The online three-dimensional bin packing process can be described as a Markov Decision Process (MDP), whose mathematical formulation is given in Equation (1).

$$M = (S, A, R, P, \gamma) \tag{1}$$

where, $S$ denotes the state set of the environment, corresponding to the observed state of the current scene; $A$ is the set of executable actions; $P$ represents the state transition probability, where $P(s_{t+1}|s_t, a_t)$ denotes the probability that the system transitions from state $s_t$ to state $s_{t+1}$ under the action $a_t$.

At each discrete time step, the agent perceives the environment to obtain the current state $s_t$, where the environmental state includes the information of the goods to be loaded and the current loading state of the container. After the agent selects and executes an action $a_t$ from the optional actions according to the policy $\pi_\theta(s_t|a_t)$, the system transitions to a new state $s_{t+1}$ with probability $P$, and the agent receives a single-step immediate reward $\gamma_t$. Based on this, the agent updates the network parameters and iterates continuously until all objects are successfully loaded into the container or the container can no longer load new objects. The core goal of this process is to learn an optimal policy $\pi_\theta$ to maximize the cumulative reward.

First, the state space, action space, and reward function are constructed. The state $s_t$ in the state space ($S$) contains the historical information required for decision-making to satisfy the Markov property, which is defined as follows:

$$s_t = (C_t, o_t) \tag{2}$$

Herein, $C_t$ denotes the space occupancy state of the container at time at $t$, which is characterized by a list of placed objects: $C_t = \{obj_1, obj_2, \dots, obj_k\}$. Each $obj_i = (x_i, y_i, z_i, l_i, w_i, h_i, \theta_i)$ includes the bottom-left corner coordinates $(x_i, y_i, z_i)$, dimensions $(l_i, w_i, h_i)$, and rotation attitude $\theta_i$ of the object; $o_t$ represents the item to be placed at time $t$, whose geometric features are consistent with those of the placed items and is characterized as $o_t = (l_o, w_o, h_o, \theta_o)$, where $\theta_o$ is the candidate rotation attitude of the item to be placed, and $(l_o, w_o, h_o)$ is the dimension information of the item to be placed.

The action space $a_t$ corresponds to the loading decision of the item to be placed $o_t$, whose core is to determine the placement position and rotation attitude of the item in the container, and it must satisfy the physical constraints of "no out-of-bounds and no overlap", which is defined as follows:

$$a_t = ((x_t, y_t, z_t), \theta_t) \tag{3}$$

Herein, $(x_t, y_t, z_t)$ is the 3D coordinate of the bottom-left corner of the item to be placed $o_t$ in the container, which must satisfy the container boundary constraint, i.e., the entire item is located inside the container after placement, with no part exceeding the container range. $\theta_t$ is the rotation attitude of the item to be placed $o_t$, and 3 to 6 core rotation directions are usually selected from a preset discrete attitude set to balance expressive ability and computational efficiency.

To avoid the action space explosion, heuristic rules are adopted to pre-screen feasible placement candidates, and only the placement schemes that satisfy the constraint conditions are included in the action set, so as to ensure the real-time performance and effectiveness of decision-making.

In the online loading mode, both the dimensions and quantity of subsequent goods to be loaded are unknown; therefore, it is necessary to balance immediate feasibility and long-term space utilization in dynamic decision-making. In the heuristic optimization framework of the three-dimensional bin packing problem, the three-dimensional bin packing of goods with different dimensions is essentially to efficiently solve an approximately optimal packing scheme under online loading constraints, and its core decisions can be summarized into three key factors: goods sequence, placement position, and placement attitude. The maximum subspace division strategy, the minimum leftover space strategy, the origin-nearest strategy, and the extreme corner placement strategy are the core components. These four strategies construct differentiated loading decision logics starting from four core optimization objectives, namely space utilization, layout compactness, effectiveness of candidate placement points, and space regularity. In the actual bin packing process, the current execution strategy is selected from the four types of strategies through a random selection mechanism, which not only gives full play to the unique advantages of each strategy but also avoids the scenario adaptation limitations of a single strategy through strategy diversity, providing solid support for the overall optimization of the bin packing scheme.

### 3.1 The maximum subspace division strategy

In the cargo positioning link, this paper adopts the corner-occupying placement strategy: the left-rear-bottom corner of the goods to be loaded is placed as close as possible to the origin of the three-dimensional Cartesian coordinate system to maximize space compactness and simplify the remaining space division logic. After the first goods are placed in the left-rear-bottom corner of the container, an irregular remaining space will be formed inside the container; to support subsequent iterative loading, this space is divided into three orthogonal subspaces (front, right, and upper) according to rules, and subsequent goods are placed into the subspaces in a fixed order. After each goods placement, three new subspaces are re-divided based on the current space state, and the "goods placement-space division" cycle is continuously executed until there are no goods to be loaded or the remaining goods cannot be placed into any subspace, at which point the loading process terminates.

Space division is a key operation in the bin packing algorithm, which is used to generate structurally regular maximum subspaces. Its purpose is to reduce space waste during the loading process and provide precise placement coordinates for goods. To ensure the effective utilization of the remaining space, the maximum subspace division principle is adopted. For the space generated after each placement, the volumes of the subspaces are calculated respectively according to the three space division methods shown in Figure 1, and the division scheme that can generate the largest continuous subspace is selected to avoid excessive fragmentation and reserve sufficient loading space for subsequent large-sized goods.

After each cargo placement is completed, the newly formed remaining space is calculated separately according to three preset division methods, and the division scheme that can generate the largest continuous subspace is selected to improve the space utilization rate of subsequent bin packing. After the space division is completed, the algorithm sequentially loads the goods into the available subspaces, and continues to perform division and update on the newly generated remaining space. As goods are continuously loaded, the container space will be gradually divided into a large number of small areas, and some areas can no longer be used due to their excessively small size.

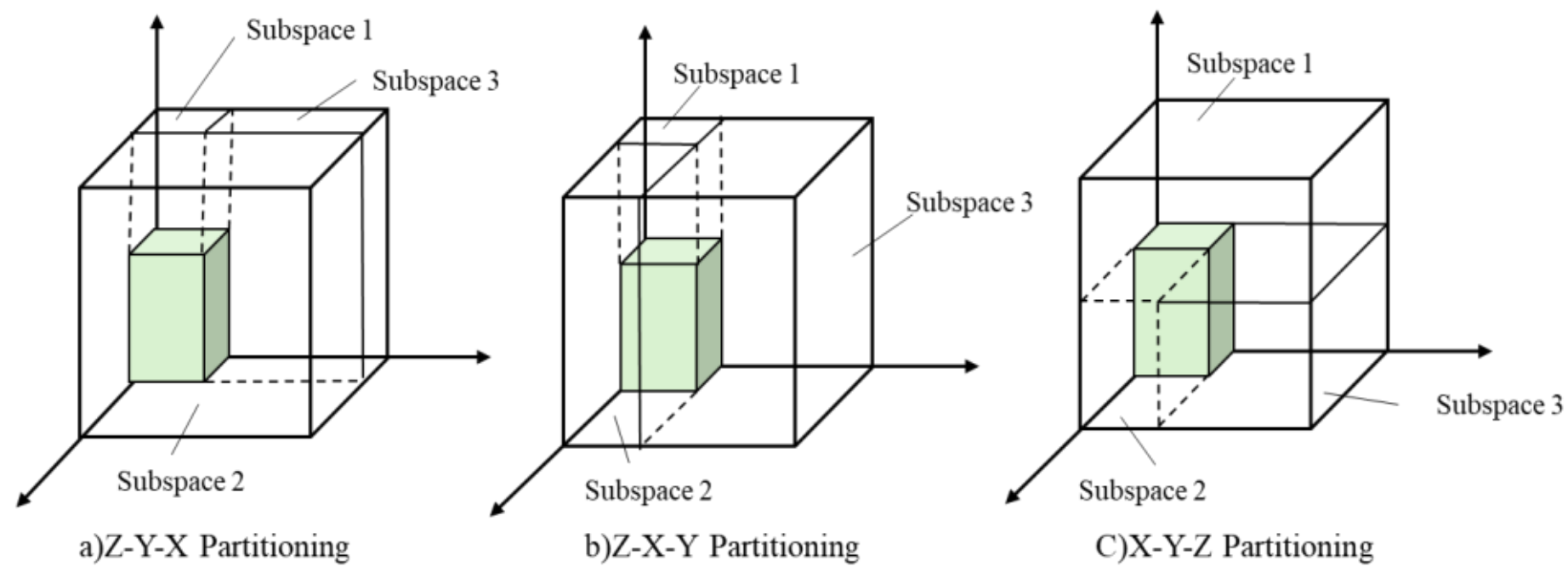

Fig. 1 Three Spatial Partitioning Methods

To alleviate the problem of space fragmentation, the algorithm introduces a space merging mechanism to re-merge adjacent and integrable free spaces into larger available spaces. Space merging follows the coplanar principle: merging can only be performed if two free spaces have a completely overlapping common contact surface; if there is no coplanar boundary, merging cannot be completed. Figure 2 shows a schematic diagram of remaining space merging.

### 3.2 The minimum leftover space strategy

The minimum leftover space strategy aims to maximize space utilization and achieve efficient bin packing by suppressing space fragmentation. Its core idea is to give priority to using the feasible space with the smallest volume and reserve large spaces for large-sized goods, thereby reducing invalid gaps.

(1) Space screening: Select all feasible spaces $S_{valid}$ that satisfy $l(s_i) \geq l,\ w(s_i) \geq w,\ h(s_i) \geq h$

(2) Space sorting: Sort the spaces in ascending order of their volume $V(s_i) = l(s_i) \cdot w(s_i) \cdot h(s_i)$.

(3) Space selection: Select $argmin_{s_i \in s_{valid}} V(s_i)$ to place the object.

This strategy is a local greedy strategy with small computational complexity and fast response, which can significantly reduce space fragmentation; however, since it only considers local optimality, it may not achieve global optimality and is usually used in combination with global search or heuristic strategies.

### 3.3 The origin-nearest strategy

The origin-nearest strategy takes the compactness of bin packing layout as its core goal. By giving priority to placing goods in the adaptive space closest to the coordinate origin, it achieves centralized placement of goods, reducing gap waste and space fragmentation. With compact layout as its core logic, this strategy makes the remaining space more regular through centralized placement, facilitating efficient loading of subsequent goods.

In the three-dimensional bin packing space, the rear-bottom-left vertex of the container is set as the coordinate origin $O = (0,0,0)$. Let the center point coordinate of the i-th remaining space be $C_i = (x_i, y_i, z_i)$. To balance computational efficiency and the actual placement characteristics of bin packing, this paper uses Manhattan distance to measure the distance between the space and the origin:

$$D_i = | x_i - 0 | + | y_i - 0 | + | z_i - 0 | = x_i + y_i + z_i \quad (4)$$

Herein, the smaller $D_i$ is, the closer the space is to the origin, and the higher its placement priority.

First, traverse all remaining spaces and screen out the set of adaptive spaces that satisfy the cargo size constraints; then calculate the Manhattan distance for all adaptive spaces, and sort them in ascending order of distance to obtain the priority sequence $D_1 \leq D_2 \leq \cdots \leq D_n$; finally, select the space with the smallest distance as the optimal placement space $S_{best} = argminS_i \in D_i$ ; after completing the placement, update the remaining space distribution and enter the next round of iterative decision-making.

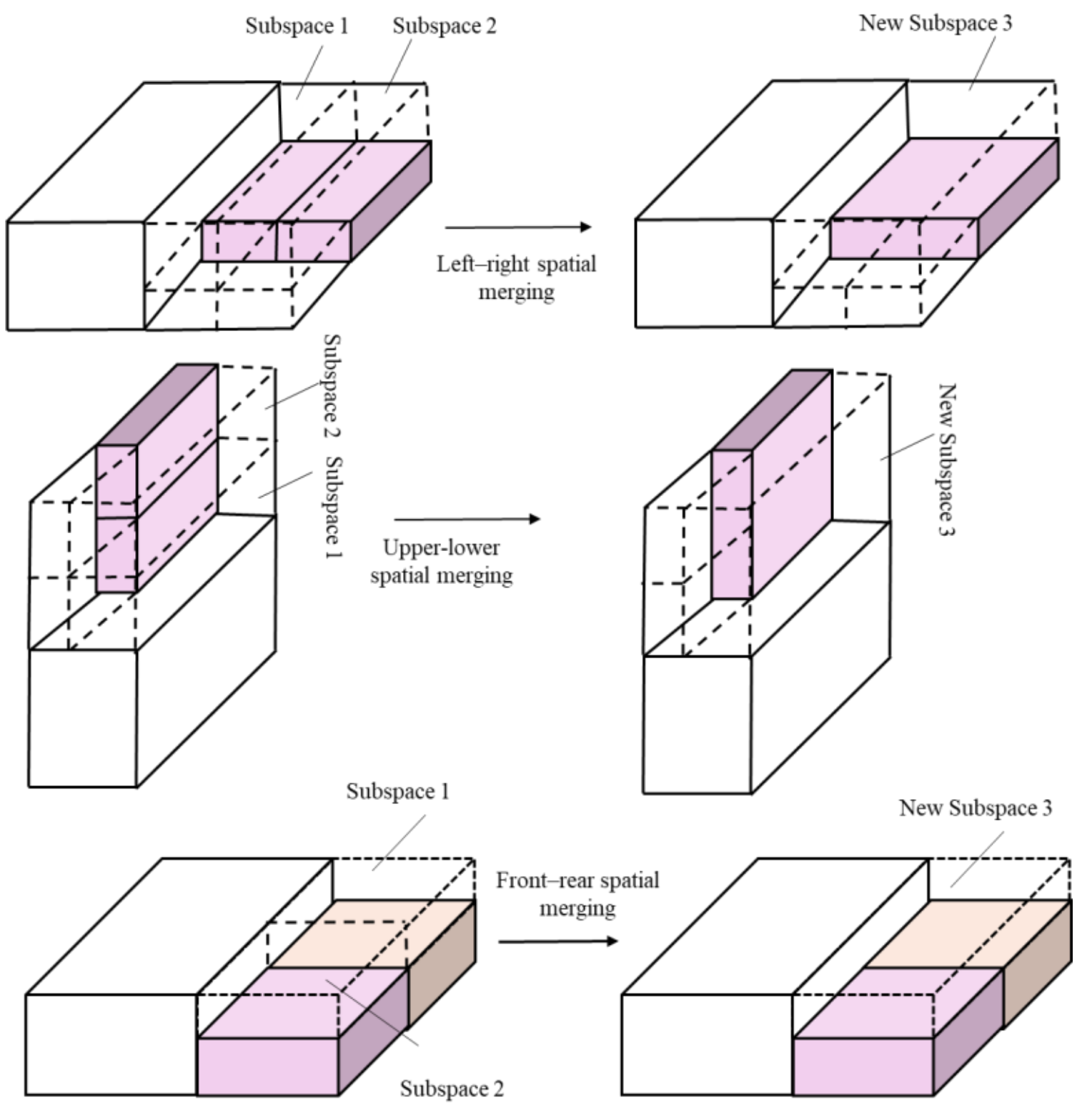

Fig. 2 Spatial Merging Strategy

### 3.4 The extreme corner placement strategy

The extreme point corner strategy identifies extreme points in the three-dimensional bin packing scenario and determines the initial candidate placement positions of goods based on these points. This achieves full utilization of corner spaces and gap spaces, while ensuring the rationality and effectiveness of candidate placement points. The selection of its extreme points is shown in Figure 3 below.

Extreme point identification is the basic link of the extreme point placement strategy, whose core task is to screen out the endpoints that are "unoccupied and can be used as the starting vertices for goods placement" from the three-dimensional bin packing space, covering two spatial scopes: the container boundary and the surface of loaded goods. The initial set of extreme points is composed of the union of the set of container boundary vertices and the set of loaded goods surface vertices, which is formally expressed as:

$$P_{init} = P_{container} \cup P_{loaded} \tag{5}$$

Herein, the set of container boundary vertices $\boldsymbol{P_{container}}$ refers to the 8 inherent corner points of the bin packing container, whose mathematical definition is as follows:

$$P_{container} = \{(x, y, z) \mid x \in \{0, L\}, y \in \{0, W\}, z \in \{0, H\}\} \tag{6}$$

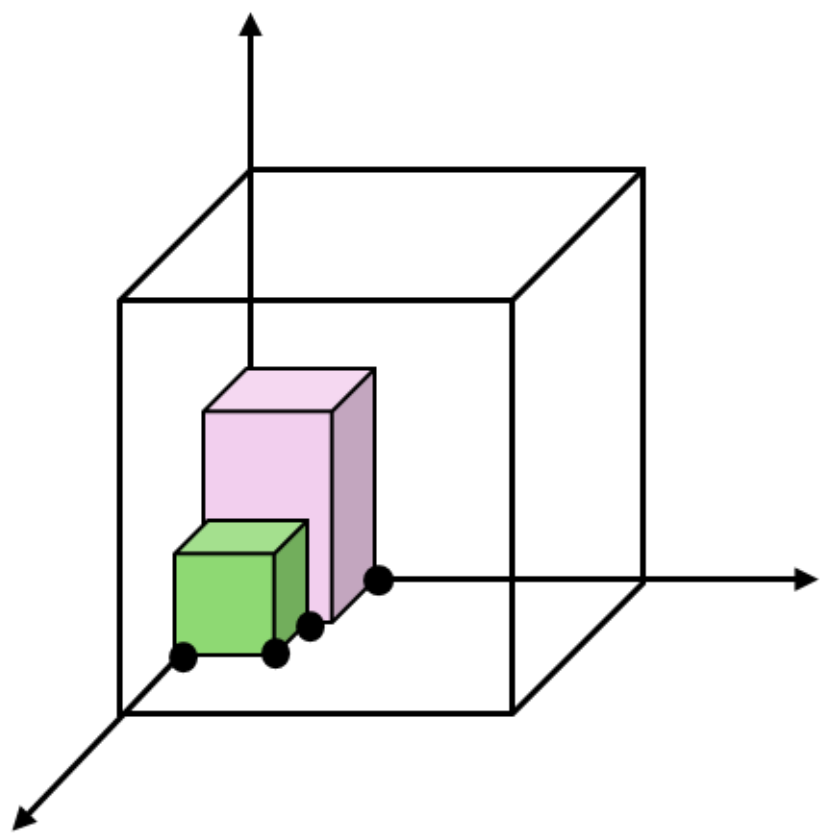

Fig. 3 Schematic Diagram of Extreme Points

The set of surface vertices of loaded goods $P_{loaded}$ refers to the vertices at the junction between all loaded goods and the remaining space, which is composed of the 8 vertices of each loaded goods. The specific expression is as follows:

$$P_{loaded} = \cup_{gj} \in G_{loaded}\{(x_j + l_j \cdot a, y_j + w_j \cdot b, z_j + h_j \cdot c) \mid a, b, c \in \{0,1\}\} \tag{7}$$

Among them, $(x_j, y_j, z_j)$ is the initial placement coordinate of the loaded goods $g_j$, and $a, b, c$ are 0-1 variables, which are used to characterize the offset direction of the goods' vertices along each coordinate axis.

The initial set of extreme points needs to go through dual constraint screening to eliminate endpoints that do not meet the placement requirements, and finally form the initial candidate extreme point set $\boldsymbol{P_{candidate}}$, which provides a basic candidate point set for subsequent links.

First, an extreme point must lie outside the spatial bounding box of any loaded item. That is, there exists no loaded item $g_j$ that can completely cover this endpoint. The mathematical constraint is given by:

$$\begin{cases} \forall p_k = (x_k, y_k, z_k) \in P_{\text{init}} \\ \quad \nexists g_j \in G_{\text{loaded}} \\ \text{s.t.} \begin{cases} x_j \le x_k \le x_j + l_j \\ y_j \le y_k \le y_j + w_j \\ z_j \le z_k \le z_j + h_j \end{cases} \end{cases} \tag{8}$$

The extreme points must also satisfy that when taking them as the starting point to place the goods $g_i$, the goods do not exceed the container boundaries, that is, the coordinates of the goods in all dimensions do not exceed the maximum boundaries of the container. The constraint expression is as

follows:

$$\forall p_k = (x_k, y_k, z_k) \in P_{\text{init}}, \begin{cases} x_k + l_i \le L \\ y_k + w_i \le W \\ z_k + h_i \le H \end{cases} \quad (8)$$

When the goods are placed at a certain extreme point, their placement behavior along the three-dimensional coordinate axes will generate new spatial endpoints. These endpoints become potential candidate points for the placement of subsequent goods. The core of the new extreme point generation link is to quantitatively generate new endpoints and update the set of extreme points.

Note: Assuming that goods $g_i$ with dimensions $l_i \times w_i \times h_i$ are placed at the extreme point $p_k = (x_k, y_k, z_k)$, the goods will occupy the corresponding space along the three directions of the x-axis, y-axis, and z-axis. The ends of the goods in each direction will generate 3 new potential extreme points. The generation formulas of the new extreme points are as follows:

$$P_{\text{new}} = \{(x_k + l_i, y_k, z_k), (x_k, y_k + w_i, z_k), (x_k, y_k, z_k + h_i)\} \quad (9)$$

The updated set of extreme points is the union of the current set of extreme points and the newly generated set of extreme points, that is:

$$P_{\text{updated}} = P_{\text{current}} \cup P_{\text{new}} \quad (10)$$

In the formula, $P_{current}$ is the set of extreme points before placing the goods, and the updated $P_{updated}$ will serve as the basic set of extreme points for the subsequent packing process, realizing the dynamic iteration of the extreme point set.

Invalid point elimination is a key link to ensure the effectiveness of the extreme point placement strategy. Its purpose is to eliminate redundant or invalid endpoints in the initial candidate set that cannot normally place goods, so as to avoid loading failure and space waste caused by the selection of invalid points. Combined with the actual characteristics of the packing scenario, three types of core invalid points need to be eliminated. First, if an extreme point is located at the corner of the container and placing the goods $g_i$ with this point as the starting point will exceed the container boundary, it is determined as an invalid point at the corner. The extreme points $p_k$ that meet the above conditions are removed from $P_{candidate}$, and this set of invalid points is denoted as $P_{corner}$. There may be duplicate points with completely consistent coordinates in the initial candidate extreme point set. These duplicate points will increase the computational burden of subsequent decisions and need to be eliminated. Let the set of duplicate points after elimination be $P_{duplicate}$. This link can effectively reduce the redundancy of the extreme point set and improve computational efficiency. If an extreme point is blocked by loaded goods or the container boundary, making it impossible to normally place goods and expand space, it is determined as an occluded invalid point. The extreme points $p_k$ that meet the above conditions are removed from $P_{candidate}$, and this set of invalid points is denoted as $P_{occluded}$.

After eliminating the above three types of invalid points, the finally obtained set of effective extreme points is the result of removing all invalid points from the initial candidate set, which is formally expressed as:

$$P_{\text{valid}} = P_{\text{candidate}} - \left(P_{\text{conner}} \cup P_{\text{duplicate}} \cup P_{\text{occluded}}\right) \quad (11)$$

$P_{valid}$ is the final candidate position set for goods placement in the extreme point placement strategy, which ensures the rationality and effectiveness of the candidate points.

# 4 Spatial Heuristic Strategy based Diffusion Reinforcement Learning Algorithm

The diffusion reinforcement learning framework based on spatial heuristic strategy is shown in Figure 4. It is divided into three core stages: heuristic pre-optimization, graph neural network feature extraction, and diffusion reinforcement learning. First, heuristic strategies such as maximum subspace segmentation and minimum remaining space are used to quickly filter the initial packing schemes, generate high-quality candidate layouts, and complete data preprocessing and optimization. Second, the packing space and cargo relationship are modeled as a graph structure. Local spatial features are aggregated through graph convolutional layers, and compact global state features are output via feature extraction and graph transformation, which effectively capture the topological and uniformity information of the packing layout. Finally, taking the output of the graph network as the state input, combined with the forward diffusion and reverse denoising mechanisms of the diffusion model, the reinforcement learning policy network learns the optimal placement action. The environmental feedback integrates the composite reward of space utilization and flatness, guiding the diffusion model to generate regular and efficient final packing schemes.

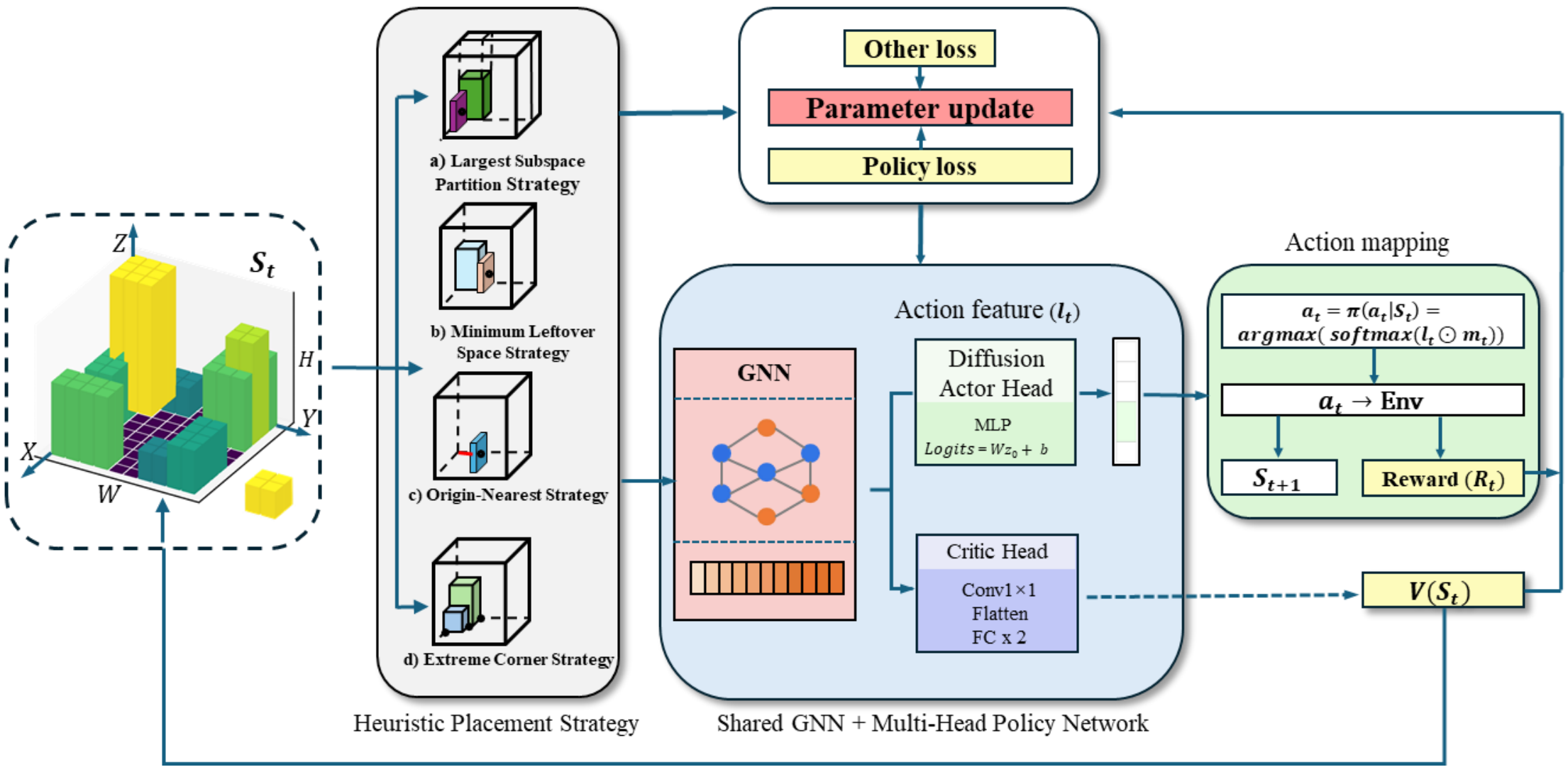

Fig. 4 Framework of Diffusion Reinforcement Algorithm Integrating Spatial Heuristic Strategy

## 4.1 Graph neural network module

The Graph Neural Network feature processing module serves as a bridge between heuristic candidate layouts and reinforcement learning decisions. It abstracts the complex 3D packing scenario into graph-structured data and efficiently models the topological dependencies between packing space and cargoes via graph convolution, global feature extraction and dynamic graph transformation, providing compact and semantically meaningful state representations for reinforcement learning.

To convert the high-dimensional and unstructured packing state into a form suitable for graph neural network processing, this paper models the state $s_t = (C_t, o_t)$ as an undirected graph $G = (V, A)$, as shown in Figure 5, where $V$ is the node set, $A \in \mathbb{R}^{N\times N}$ is the adjacency matrix, and $N$ is the total number of nodes. Together, they encode all core information of the packing layout. The node set $V$ contains two core types of nodes, which respectively map the two major components of the state space to ensure that all historical information required for decision-making is retained.

Placed object nodes are denoted as $C_t = \{obj_1, obj_2, \ldots, obj_k\}$. The feature vector of each node encodes all attributes of the corresponding object $obj_i = (x_i, y_i, z_i, l_i, w_i, h_i, \theta_i)$, including the coordinates of the lower-left corner, 3D dimensions, and rotation posture, fully characterizing the spatial occupancy state of the container. The node of the item to be placed matches $o_t$ precisely. Its feature vector encodes the geometric attributes of the item to be placed $o_t = (l_o, w_o, h_o, \theta_o)$, covering dimension information and candidate rotation postures, ensuring that the decision-making features of the item to be placed are fully embedded in the graph structure.

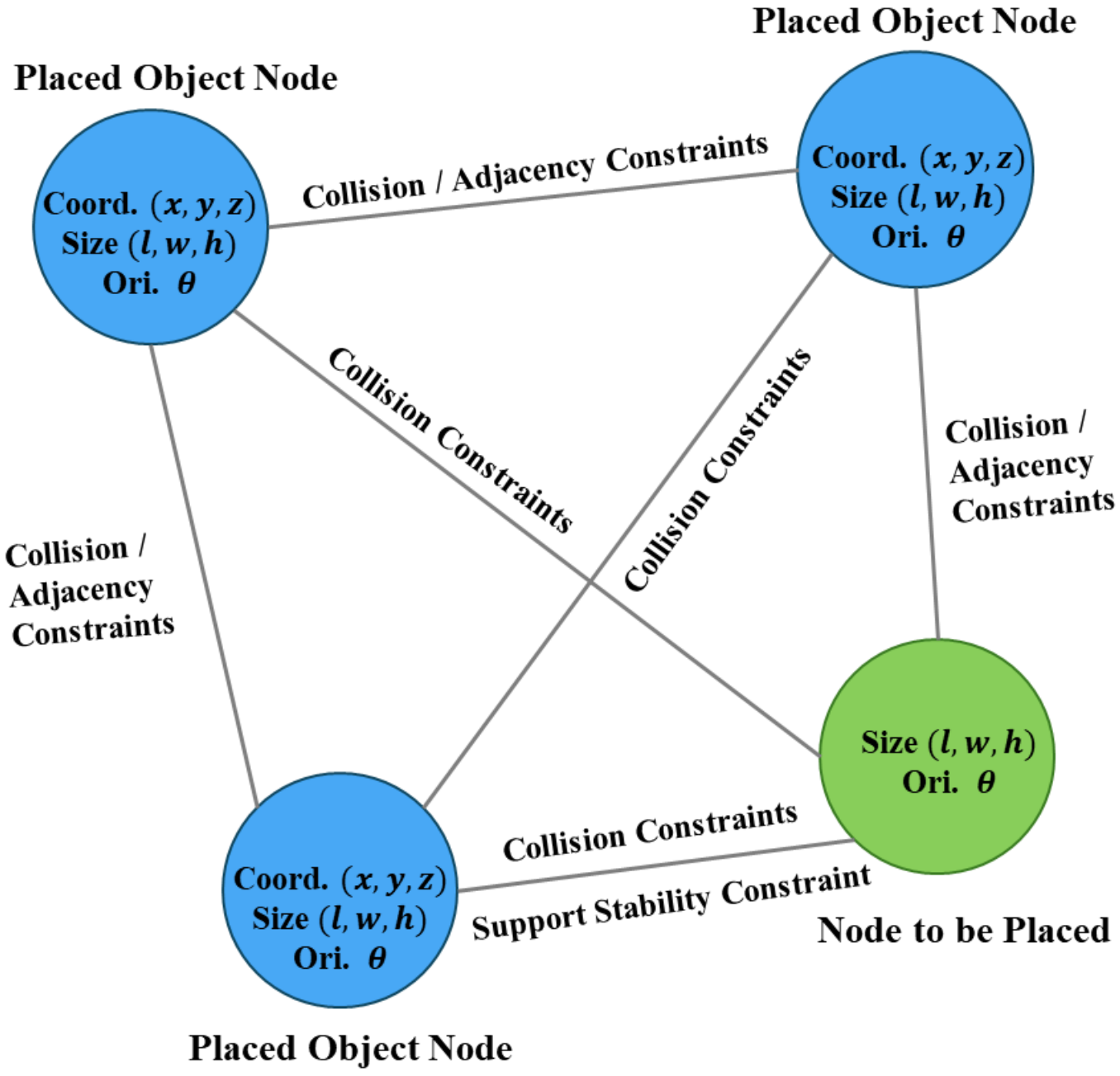

Fig. 5 Graph Structure Modeling of Packing State

The adjacency matrix $A$ is the matrix representation of the edge set $E$, used to quantify the spatial constraint relationships between nodes. Its block structure and physical meaning are as follows:

$$A = \begin{bmatrix} A_{11} & A_{12} \\ A_{21} & A_{22} \end{bmatrix} \tag{12}$$

where rows and columns both correspond to placed object nodes. $A_{11}[i, j] = w_{ij}$ represents the spatial constraint strength between the $i$-th and $j$-th placed objects, and is 0 if there is no constraint.

$A_{12} \in \mathbb{R}^{k \times 1}$: rows correspond to placed object nodes, column corresponds to the object to be placed. $A_{12}[i, 1] = w_i$ denotes the potential collision risk and position compatibility between the $i$-th placed object and the object to be placed. $A_{21} = A_{12}^T$: symmetry holds due to the undirected graph nature, representing the reverse spatial constraint from the object to be placed to the placed objects. $A_{22} = 0$: only one object is placed at each step; there is no self-constraint for the node of the object to be placed, so its value is 0.

The core role of the adjacency matrix is to define the information propagation path for graph convolution. Feature aggregation is only performed between nodes with non-zero connections, ensuring efficient transmission of spatial constraint information. Meanwhile, the weight $w_{ij}$ quantifies the constraint strength, realizing fine-grained encoding of topological relationships.

Graph convolutional layers are the core of local spatial feature extraction, used to fuse node self-attributes and spatial constraint information encoded by the adjacency matrix, as shown in Figure 6. This paper adopts a graph convolutional network to implement feature aggregation, and the feature update formula of node $\boldsymbol{i}$ at layer $\boldsymbol{l+1}$ is:

$$\boldsymbol{h}_i^{(l+1)} = \sigma\left(\boldsymbol{W}^{(l)} \sum_{j \in \mathcal{N}(i) \cup \{i\}} \frac{A_{ij} \cdot \boldsymbol{h}_j^{(l)}}{\sqrt{|\mathcal{N}(i)| \cdot |\mathcal{N}(j)|}} + \boldsymbol{b}^{(l)}\right) \tag{13}$$

where $h_i^{(l)}$ is the feature vector of node $i$ at the $l$-th layer; $N(i)$ is the set of neighbor nodes of node $i$; $W^{(l)}$ and $b^{(l)}$ are learnable weights and biases; $\sigma$ is the nonlinear activation function that enhances feature representation ability.

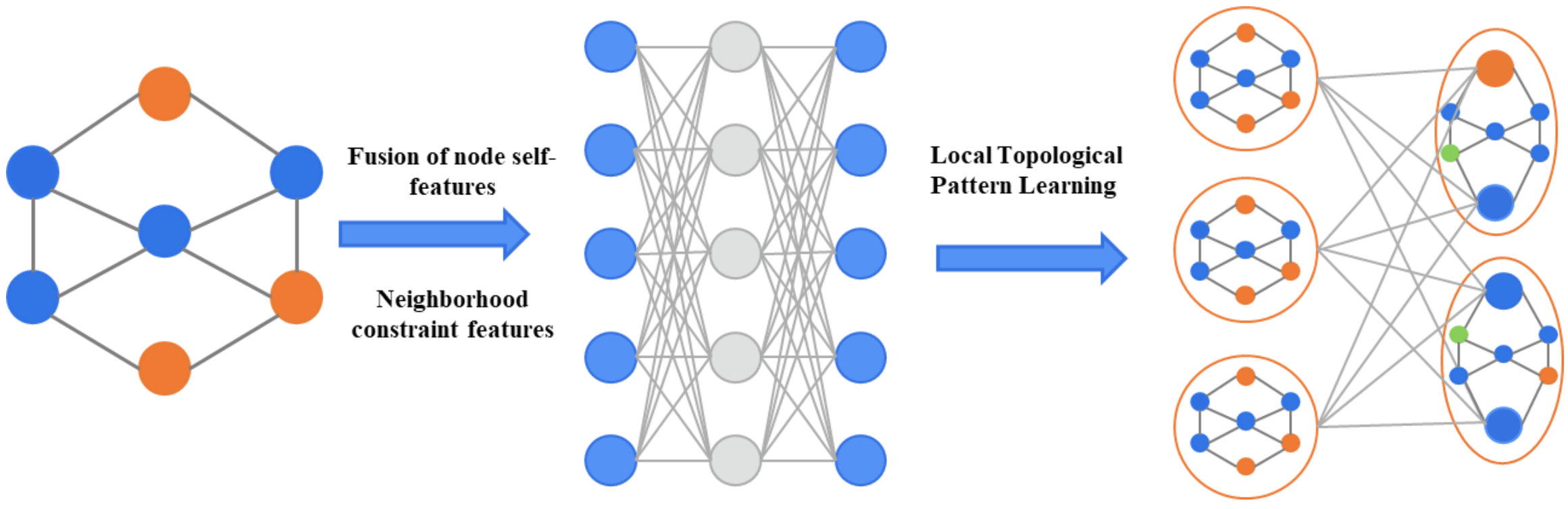

Fig. 6 Local Feature Aggregation via Graph Convolution

On the basis of multi-layer graph convolution, the high-dimensional graph features are compressed into a fixed-dimensional global state vector through global average pooling. The process is shown in Figure 7, which serves as the final state input for reinforcement learning:

$$s = \frac{1}{N} \sum_{i=1}^{N} h_i^{(L)} \tag{14}$$

where $L$ is the total number of graph convolution layers, $h_i^{(L)}$ is the node feature output by the last graph convolution layer, and $s$ is the state feature vector with fixed dimension.

This vector condenses all the core decision-making information of the packing layout: it not only includes the spatial occupancy state of the placed objects and the geometric features of the objects to be placed, but also contains the spatial constraints between objects, position compatibility and height alignment of the original state. It meets the requirement of state completeness for Markov property, and avoids the dimension explosion problem caused by high-dimensional original states.

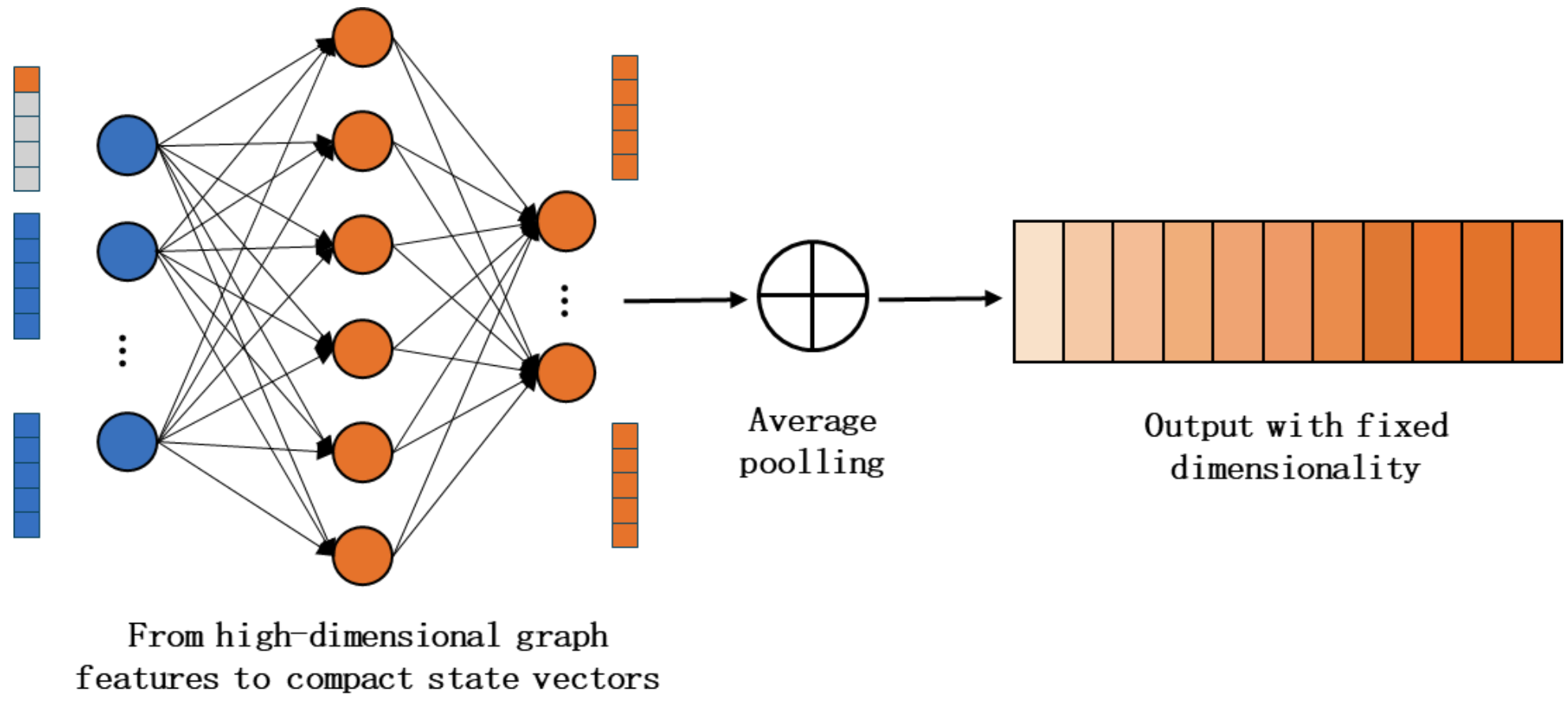

Fig. 7 Global Feature Extraction and State Encoding

### 4.2 Diffusion reinforcement learning

In the online three-dimensional bin packing problem, the high dimensionality, multimodal distribution of the action space, and the instability of the strategy are major challenges faced by current research. This paper adopts the diffusion reinforcement learning algorithm as the reinforcement learning optimization framework, which not only ensures training stability but also improves the strategy's search and expression capabilities in the continuous action space.

The diffusion model defines the data generation process as a gradual denoising generation process [66], which can effectively fit the complex action distribution structure. By converting the reinforcement learning problem into a conditional sampling problem, it generates the optimal packing actions matching the current environmental state, thereby improving the performance of the robot's packing strategy in a dynamic environment. In the bin packing problem, the diffusion model generates a reasonable packing action distribution by gradually removing noise. This process can not only capture the complex multimodal distribution in the action space but also dynamically adjust the action generation strategy according to the current environmental state. Specifically, the diffusion model models the generation process of packing actions as an iterative denoising process from random noise to effective actions: first, random noise is injected into the action space at the initial moment to break the limitation of a single decision-making mode; then, in each denoising step, the model takes the current environmental state features as conditions, predicts and eliminates noise components through a neural network, and gradually approaches the action distribution that meets physical constraints and optimization objectives. The main process is shown in Figure 8.

During the forward noising process, the action $x_t$ at the current moment is only related to $x_{t-1}$ at the previous moment, and this process can be regarded as a Markov process, which satisfies:

$$q(x_{1:T}|x_0) = \prod_{t=1}^{T} q\,(x_t|x_{t-1}) \tag{15}$$

$$q(x_t|x_{t-1}) = N(x_t\,; \sqrt{1-\beta_t x_{t-1}}\,, \beta_t I) \tag{16}$$

$q(x_t|x_{t-1})$ denotes the probability distribution of obtaining the noisy action $x_T$ after $T$ steps of noising process starting from the original action $x_0$.

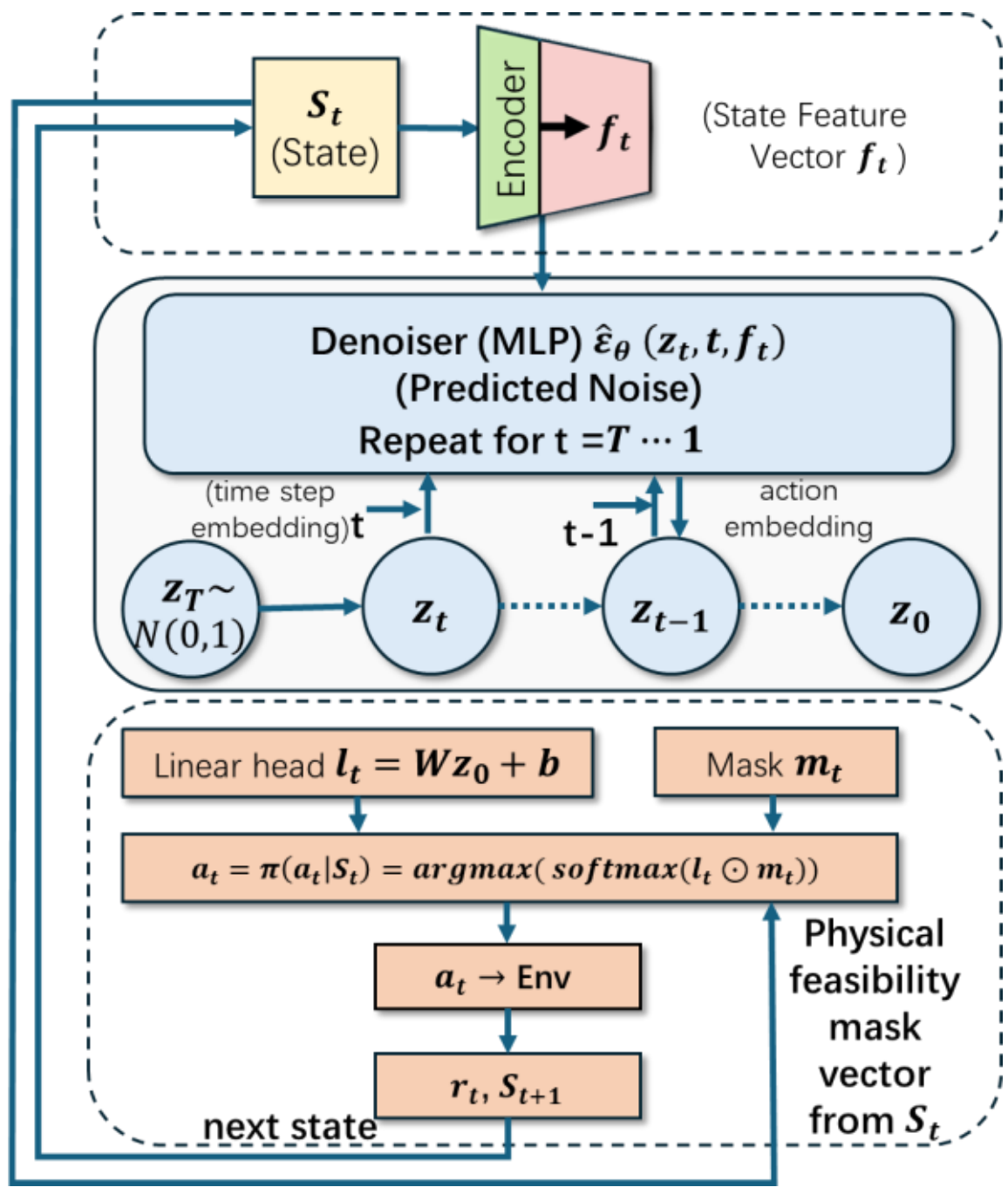

Fig. 8 Diffusion Probability Model

Each step $q(x_t|x_{t-1})$ represents the conditional probability distribution of obtaining the current image $x_t$ given the image $x_{t-1}$ from the previous step in the t-th step. This is a Gaussian distribution with a mean of $\sqrt{1-\beta_t x_{t-1}}$ and a variance of $\beta_t$, where $\beta_t$ is a predetermined noise variance coefficient and $I$ is the identity matrix. Among them, the $\beta_t$ values for different t are pre-defined, gradually increasing from time 1 to $T$ (which can be Linear, Cosine, etc.), satisfying

Based on the above formulas, $x_t$ can be obtained through reparameterization sampling, where $\epsilon \sim N(0, I)$ and $\alpha_t = 1 - \beta_t$; then $\vec{\alpha}_t = \prod_{i=1}^{T} \alpha_i$ is defined, and through derivation, the relationship between $x_t$ and $x_0$ can be obtained.

$$q(x_t|x_0) = N(x_t; \sqrt{\bar{\alpha}_t x_0}, (1-\bar{\alpha}_t)I) \tag{17}$$

During the reverse denoising process, if the reverse process $q(x_{t-1}|x_t)$ is obtained, the action probability distribution can be gradually restored through random noise $x_T$. DDPM uses a neural network $p_\theta(x_{t-1}|x_t)$ to fit the reverse process $q(x_{t-1}|x_t)$. Through the above formula (18), formula (19) can be derived.

$$q(x_{t-1}|x_t, x_0) = N(x_{t-1}|\tilde{\mu}_t(x_t, x_0), \tilde{\beta}_t I) \tag{18}$$

$$p_\theta(x_{t\text{--}1}|x_t) = N(x_{t\text{--}1}|\mu_\theta(x_t, t), \Sigma_\theta(x_t, t)) \tag{19}$$

The diffusion model fits the noise prediction model $\epsilon_\theta(\cdot)$ through a U-Net neural network, thereby indirectly obtaining the mean value $\mu_\theta$ to calculate $x_{t-1}$. The variance is usually fixed, not directly fitted through the neural network, but determined by the parameters of the diffusion process.

$$\mu_\theta = \frac{1}{\sqrt{\alpha_t}}(x_t - \frac{1-\alpha_t}{\sqrt{1-\bar{\alpha}_t}}\epsilon_{\theta(x_t,t)}) \tag{20}$$

In the above formulas, $t$ and $x_t$ are known terms, and it is only necessary to use a neural network to fit $\epsilon_{\theta(x_t,t)}$. The noise is added during the network training process.

The generation of actions is realized through conditional sampling of the diffusion model. When the model is sufficiently smooth, the reverse diffusion process can be approximated as a Gaussian distribution [29]:

$$\mu_\theta(x_t, t) = x_t + \sigma_t^2 \nabla_{x_t} \log p_\theta\,(x_t | y = 1) \tag{21}$$

This relationship provides a direct conversion between classifier-guided samplings, which is used for generating class-conditional images [30] and setting up reinforcement learning problems. First, a diffusion model is trained on the states and actions of all available trajectory data, and then a separate model is trained to predict the cumulative return of trajectory samples. By adjusting the mean μ of the reverse process, the trajectory sampling process can be guided, thereby generating the optimal packing actions.

In the packing task, the introduction of the diffusion model enables the agent to dynamically generate a reasonable distribution of packing actions based on the height map of the current packing environment and the geometric information of the goods to be packed. Through the process of gradual denoising, the diffusion model can capture the complex multi-modal distribution in the packing action space, thereby improving the quality and diversity of the strategy. Experimental results show that the reinforcement learning strategy based on the diffusion model is significantly superior to traditional methods in terms of space utilization and convergence efficiency, demonstrating good adaptability and generalization ability [14].

### 4.3 Loss function

To simultaneously optimize space utilization and layout flatness in continuous-space online 3D packing, and avoid space fragmentation caused by drastic local height fluctuations, we first construct a normalized nonlinear flatness penalty, and fuse it with the volume reward to form a composite reward function. This ensures that the penalty term has sufficient constraint strength, which can effectively guide the strategy to optimize towards a regular, uniform, and high-utilization direction.

Space utilization is the core optimization objective of 3D packing. To ensure that the reward signal matches the subsequent penalty terms in numerical scale and improve the stability of reinforcement learning training, this paper adopts the normalized volume reward as the basic reward term, confining the profit value within a reasonable range to avoid strategy update oscillation caused by excessively large or small values.

At time step $\boldsymbol{t}$, if the current goods to be placed can be successfully put into the box, the immediate utilization reward is defined as:

$$r_{vol,t} = \frac{v_t}{V_{bin}} \tag{22}$$

Among them, $v_t = s_t^x \cdot s_t^y \cdot s_t^z$ represents the volume of the current goods, where $s_t^x$,$s_t^y$,$s_t^z$ are the dimensions of the goods in the three coordinate axes respectively; $V_{bin} = S^x \cdot S^y \cdot S^z$ is the total volume of the box, and $S^x$,$S^y$,$S^z$ are the external dimensions of the box respectively. This design restricts the single-step volume reward within the range of [0,1], enabling the subsequent flatness penalty to effectively guide the strategy on the same scale. If there is no feasible placement position for the current goods, then $r_{vol,t} = 0$, and the current packing round terminates.

In the continuous packing space, the height field presents a continuous distribution characteristic, and it is impossible to conduct height statistics through discrete grids. To effectively constrain the layout flatness without damaging the spatial continuity, this paper designs a local height difference calculation method based on continuous integration, and constructs a three-stage piecewise nonlinear penalty function to achieve the effects of tolerance to small fluctuations, constraint on moderate fluctuations, and strong penalty on large fluctuations, as shown in Figure 9.

It not only avoids the unreasonable penalty of effective placement due to slight height fluctuations, but also forms sufficient restraint on the severely fragmented layout, enabling the penalty term to truly participate in the strategy decision-making.

Let the placement coordinates of the goods on the continuous plane at the bottom of the box be (x, y), and the bottom surface occupies a width of w and a length of l. Then the continuous support area of the goods in the plane can be expressed as $[x, x + w] \times [y, y + l]$. To characterize the local height level formed after the goods are placed, the area-weighted average of the continuous height field within the support area of the goods is calculated to eliminate the impact of slight height fluctuations within the area. Its calculation formula is:

$$\bar{h}_{obj} = \frac{1}{w \cdot l} \iint_{[x,x+w]\times[y,y+l]} h(u, v) du dv \tag{23}$$

Among them, $\boldsymbol{h}(\boldsymbol{u}, \boldsymbol{v})$ is the height field of the upper surface of the stacked goods at the continuous position $(\boldsymbol{u}, \boldsymbol{v})$ inside the box.

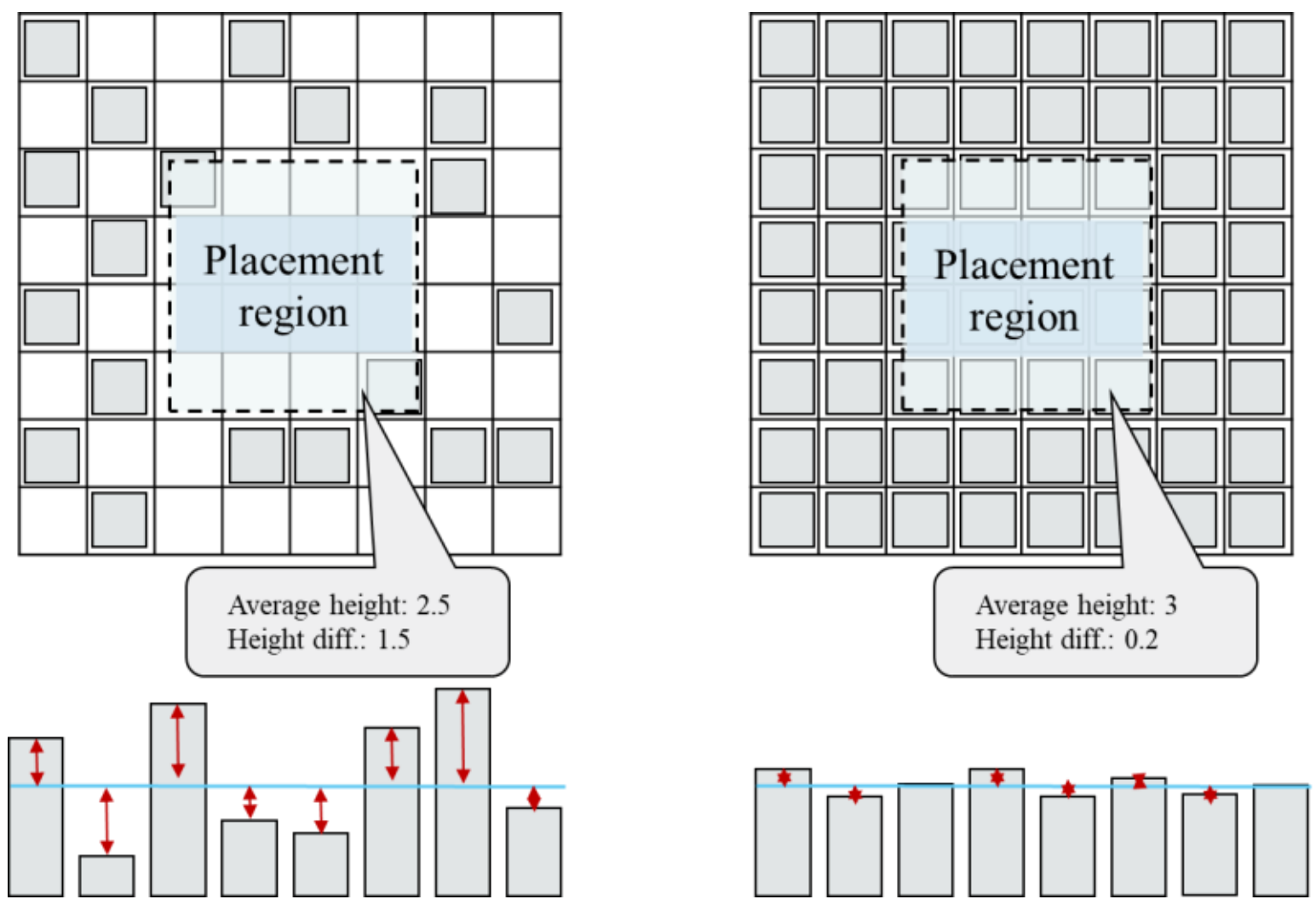

Fig. 9 Visual Comparison of Local Height Difference Penalty Mechanism for 3D Bin Packing

To objectively measure the degree of local height fluctuation, take the goods support area as the center and expand outward by a fixed width d to form a continuous annular neighborhood. After excluding the area occupied by the goods themselves, this neighborhood can be expressed as:

$$\Omega_{neigh} = [x - d, x + w + d] \times [y - d, y + l + d]/[x, x + w] \times [y, y + l] \tag{24}$$

The average height of the neighborhood reflects the overall height level of the area around the goods, serving as a reference benchmark for height fluctuations. Its calculation formula is:

$$\bar{h}_{neight} = \frac{1}{A_{neight}} \iint_{\Omega_{neight}} h(u, v) du dv \tag{25}$$

Among them, $\boldsymbol{A_{neigh}}$ is the total area of the neighborhood region, which is used to normalize and average the neighborhood height field.

To eliminate the impact of box size and cargo specifications on the absolute value of height difference, and to ensure the consistency of the penalty mechanism in different scenarios, the local height difference is normalized as follows:

$$\Delta h_{norm} = \frac{|h_{obj} - h_{height}|}{H_{bin}} \tag{26}$$

Among them, $H_{bin}$ is the total height of the box, $|\cdot|$ is the absolute value operation, which is used to eliminate the impact of the height direction and only retain the information of the undulation amplitude. After normalization, $\Delta h_{norm} \in [0,1]$, ensuring that the penalty function has a stable constraint effect under different box configurations.

To achieve refined constraints, this paper designs a three-stage piecewise penalty function, which exerts differentiated penalty intensity on different degrees of height undulation, balancing layout flexibility and regularity:

$$r_{\text{flat},t} = \begin{cases} 0 & \Delta h_{\text{norm}} \leq \alpha \\ -\lambda_1 \cdot \Delta h_{\text{norm}} & \alpha < \Delta h_{\text{norm}} \leq \beta \\ -\lambda_2 \cdot (\Delta h_{\text{norm}})^2 - C & \Delta h_{\text{norm}} > \beta \end{cases} \tag{27}$$

Among them, $\alpha = 0.05$ is the no-penalty threshold; when the height fluctuation is lower than 5% of the box height, it is regarded as a reasonable layout without imposing penalties. $\beta = 0.30$ is the switching threshold for strong penalties; if the height fluctuation is lower than 5% of the box height, it is considered a reasonable layout without imposing penalties. $\lambda_1 = 0.5$ is the linear penalty coefficient for moderate undulation, exerting mild constraints on slight excessive undulation; $\lambda_2 = 2.0$ is the square penalty coefficient for significant undulation, making the penalty value rise rapidly with the height difference, forming strong inhibition; $C = 0.05$ is the fixed penalty bias, further enhancing the penalty intensity for severely fragmented layouts.

This piecewise mechanism ensures that the penalty term is at the same order of magnitude as the volume reward, which can significantly affect the ranking of action scores, enabling the strategy to actively abandon placement schemes with excessively large height undulation and prioritize flat and regular layouts. To achieve refined constraints, this paper designs a three-stage piecewise penalty function, which exerts differentiated penalty intensity on different degrees of height undulation, balancing layout flexibility and regularity:

This piecewise mechanism ensures that the penalty term is at the same order of magnitude as the volume reward, which can significantly affect the ranking of action scores, enabling the strategy to actively abandon placement schemes with excessively large height undulation and prioritize flat and regular layouts.

Combine the normalized volume reward with the piecewise flatness penalty by weighting to obtain the final immediate reward for a single step:

$$r_t = r_{vol,t} + r_{flat,t} \tag{28}$$

The composite reward function takes space utilization as the core optimization objective, and at the same time restricts the flatness of the packing height in a perceptible and adjustable way, suppressing the fragmented distribution in the continuous space. It enables the agent to continuously optimize the placement strategy in the continuous action space, and finally achieves the online 3D packing effect with high space utilization, high layout regularity and high subsequent compatibility.

The optimization objective of the reinforcement learning model is to maximize the cumulative discounted reward, and the overall loss consists of the joint composition of the policy network loss and the value network loss. First, define the advantage function, which is used to measure the profit advantage of the current action compared with the average state value:

$$A_t = r_t + \gamma V(s_{t+1}) - V(s_t) \tag{29}$$

Among them, $\boldsymbol{r_t}$ is the composite immediate reward designed in this paper, $\boldsymbol{\gamma = 1}$ is the discount factor, and $\boldsymbol{V(s_t)}$ is the long-term cumulative reward estimation of the state s_t by the value network.

The policy network takes the state characteristics of the continuous packing space as input and outputs the probability distribution of the goods placement position. Its optimization objective is to maximize the expected cumulative reward. The policy gradient method is adopted to construct the loss function, converting the maximization problem into a minimization objective solvable by gradient descent:

$$L_{actor} = -\frac{1}{T}\sum_{t=1}^{T} A_t \cdot log\pi\theta(a_t \mid s_t) \tag{30}$$

Among them, $\boldsymbol{\pi_\theta(a_t \mid s_t)}$ is the action probability output by the policy network; $\boldsymbol{T}$ is the effective time step length of the current packing round; the negative sign realizes the conversion from the maximization objective to the minimization loss.

The value network is used to fit the true cumulative discounted reward of the state, providing a stable and low-variance value benchmark for policy gradient calculation. The mean squared error is used to measure the deviation between the value estimate and the actual reward, and the value loss is constructed as follows:

$$L_{critic} = \frac{1}{T}\sum_{t=1}^{T}(r_t + \gamma V(s_{t+1}) - V(s_t))2 \tag{31}$$

To synchronously optimize the ability of strategy decision-making and value evaluation, the strategy loss and value loss are directly weighted and summed to obtain the final jointly optimized total loss function:

$$L_{total} = L_{actor} + L_{critic} \tag{32}$$

During the training process, the model takes minimizing the total loss as the goal, continuously updates the parameters of the policy network and the value network, enabling the agent to gradually converge to an online 3D packing strategy that balances high utilization, high regularity and strong generalization in the continuous packing space.

## 5 Experiments

In this experiment, the RS, CUT-1, and CUT-2 datasets proposed by Zhao [20] were used to train and test the diffusion reinforcement learning model. These datasets were designed to evaluate the performance of algorithms for the online three-dimensional BPP. The RS dataset was generated through random sampling of item sequences without specific order. It was used to test the algorithm’s robustness and efficiency in random scenarios. The CUT-1 dataset was created by dividing boxes into items and sorting them by Z-coordinate. It was suitable for testing the algorithm's ability to handle structured data. The CUT-2 dataset was developed by considering stacking dependencies among items and generating sequences based on dependency order. The core training hyperparameters and key execution parameters of the algorithm are shown in Table 1.

The construction of training and test datasets based on public datasets follows a unified format. In the discrete solution space scenario, the length, width, and height of the container are specified as L=W=H=10. Item dimensions are positive integers ranging from 1 to 5, satisfying the constraints: $1 \leq l_i \leq \frac{L}{2}, 1 \leq w_i \leq \frac{L}{2}, 1 \leq h_i \leq \frac{L}{2}$. Placement coordinates are integers, with the action space defined as: $x \in \{0,1,\ldots,10-l_i\}, y \in \{0,1,\ldots,10-w_i\}, z \in \{0,1,\ldots,10-h_i\}$. Only extreme points adjacent to the container boundary or already placed items are allowed as candidate placement positions, ensuring spatial compactness and constraint compliance.

In the continuous solution space scenario, the container is normalized to L=W=H=1，to adapt to continuous coordinate optimization in deep reinforcement learning. Item dimensions are sampled from the uniform distribution $s^d \sim U(0.1,0.5)$, satisfying: $0.1 \leq l_i \leq 0.5, 0.1 \leq w_i \leq 0.5, 0.1 \leq \hbar_i \leq 0.5$. Placement coordinates are arbitrary real numbers, with the action space: $x \in [0,1-l_i], y \in [0,1-w_i], z \in [0,1-\hbar_i]$. Additional complex constraints such as center-of-gravity balance and stability must be satisfied, which is closer to the continuous optimization requirements of real industrial scenarios.

Tab. 1 Hyperparameter Settings

| Parameter | Value | Description |
|---|---|---|
| $\gamma$ | 1.0 | discount factor |
| $d_m$ | 64 | Input embedding dimension |
| $n_h$ | Adaptive | No. of attention heads |
| $n_l$ | 1 | No. of GAT layers |
| $d_{ll}$ | 128 | Hidden layer dimension of feedforward network |
| $g_c$ | 0.5 | Maximum gradient clipping |
| $n_p$ | 64 | Number of training parallel processes |
| $n_s$ | 5 | n-step sampling length |

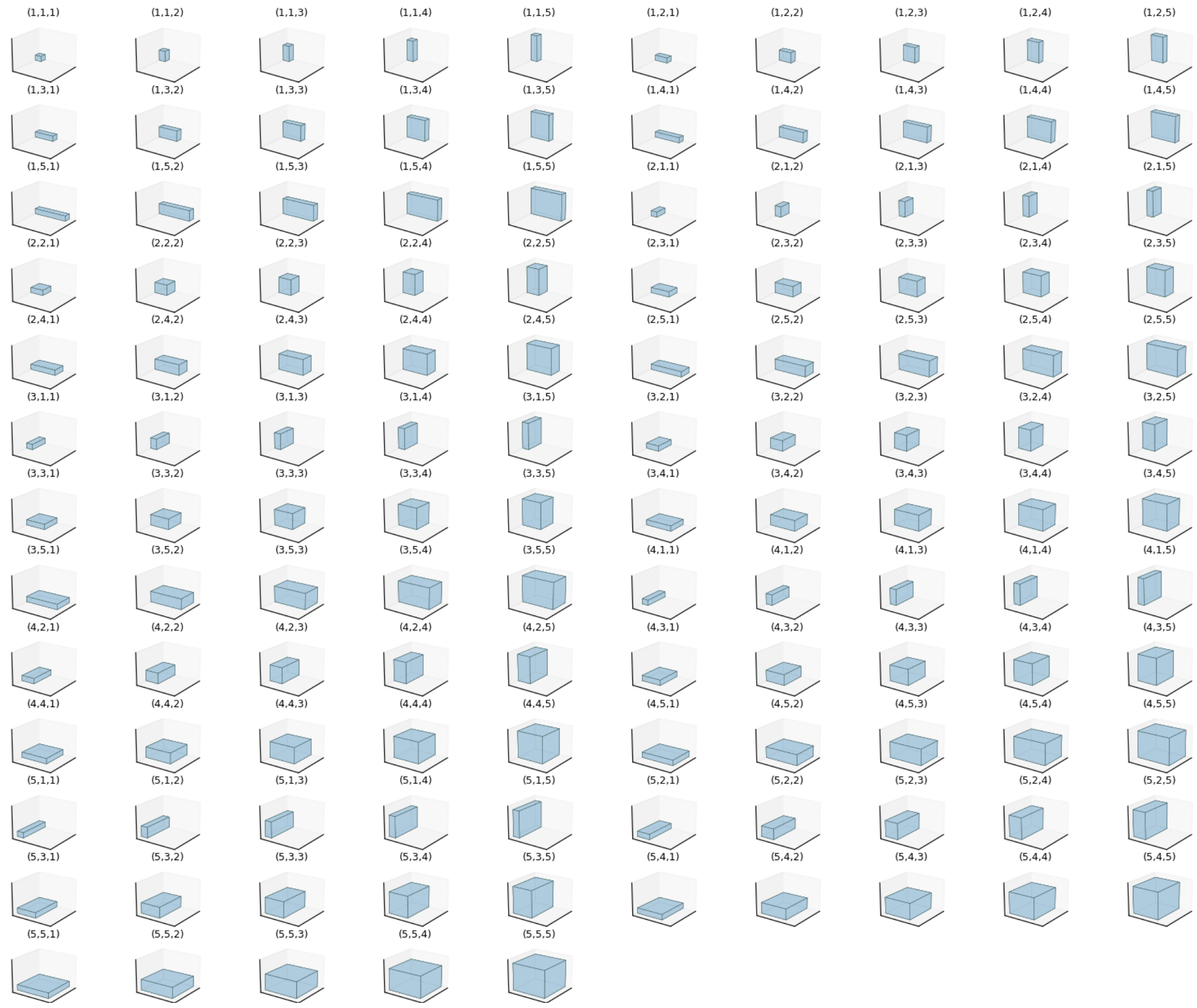

Fig. 10 Schematic Diagram of Box Configurations with Size Ranging from 1 to 5

Three different test settings Setting1, Setting2 and Setting3 are adopted as Table 2 [21]. For setting 1, The box cutting method is adopted, and objects are sorted according to stacking dependencies. Objects are only allowed to rotate along two horizontal directions, resulting in better controllability. For setting 2, A random sampling sequence is adopted, allowing objects to rotate along six directions. It has higher degrees of freedom and is closer to actual scenarios. For setting 3, Inherits the sequence method of Setting 2, retains only two horizontal rotation directions, and attaches density information following a 0~1 distribution to each object, introducing additional randomness to test the model's adaptability.

Tab. 2 Comparison of Characteristics of Different Datasets

| Setting | Constraints | Numbers | Object properties | Test objectives |
|---|---|---|---|---|
| Setting1 | Basic Constraints + Placement stability | 2 | No extra properties | Stability, robot-friendliness |
| Setting2 | Only basic non-overlapping / non-crossing constraints | 6 | No extra properties | Only space utilization |
| Setting3 | Basic constraints + stability + density | 2 | With ρ~(0,1] | Complex constraints, load balancing |

## 5.1 comparison of algorithms

To verify the effectiveness and superiority of the proposed OURS method for the 3D packing problem, Figure 11 presents the convergence curves of our method compared with various PCT-series baseline algorithms under three typical experimental settings. From the overall trend, the proposed diffusion reinforcement learning method achieves the best performance in three key dimensions: convergence speed, final performance, and training stability, and significantly outperforms existing baseline methods.

In terms of convergence speed, the proposed method exhibits extremely fast fitting ability under all three settings. It rises rapidly and enters a stable convergence stage at the early stage of iteration, and achieves a significant performance gap over other methods in the early training phase. It can reach higher space utilization with fewer iterations, greatly improving training efficiency. In terms of final convergence performance, the curve of the proposed method is stably at the top under all settings. The final packing utilization is significantly higher than that of other comparison algorithms, and optimal indicators are achieved on all three datasets, fully demonstrating stronger spatial layout optimization capability. In terms of convergence stability, the curve of the proposed method shows extremely small fluctuations after convergence, with no obvious oscillation or drop, indicating a more stable policy learning process and more reliable decision output.

The above comparison results fully demonstrate that the proposed method can not only achieve faster training convergence speed and higher packing space utilization, but also possess stronger robustness, generalization and scene adaptability in complex and variable packing scenarios, which effectively verifies the advancement and practicability of the proposed algorithm for the 3D packing problem.

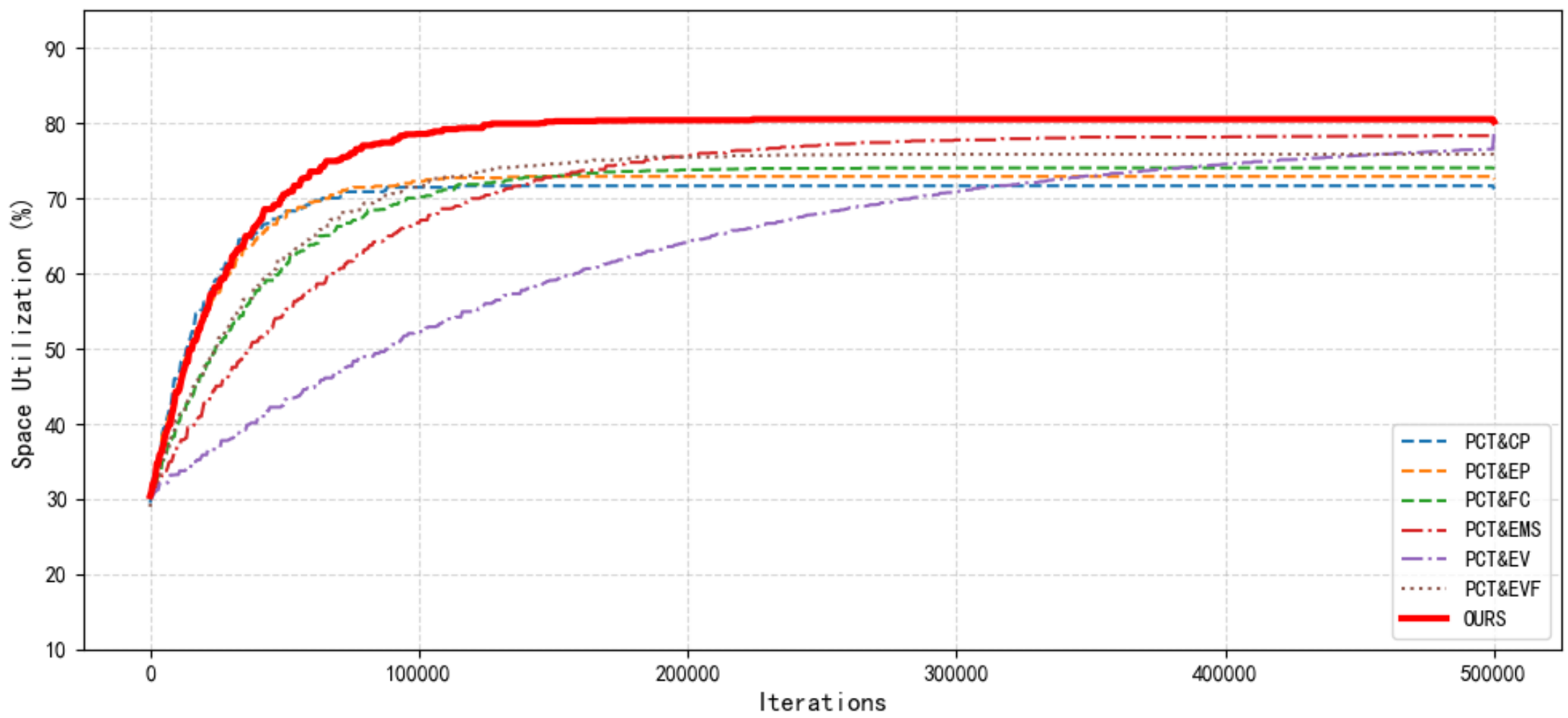

(a) Convergence curves of space utilization under Setting 1

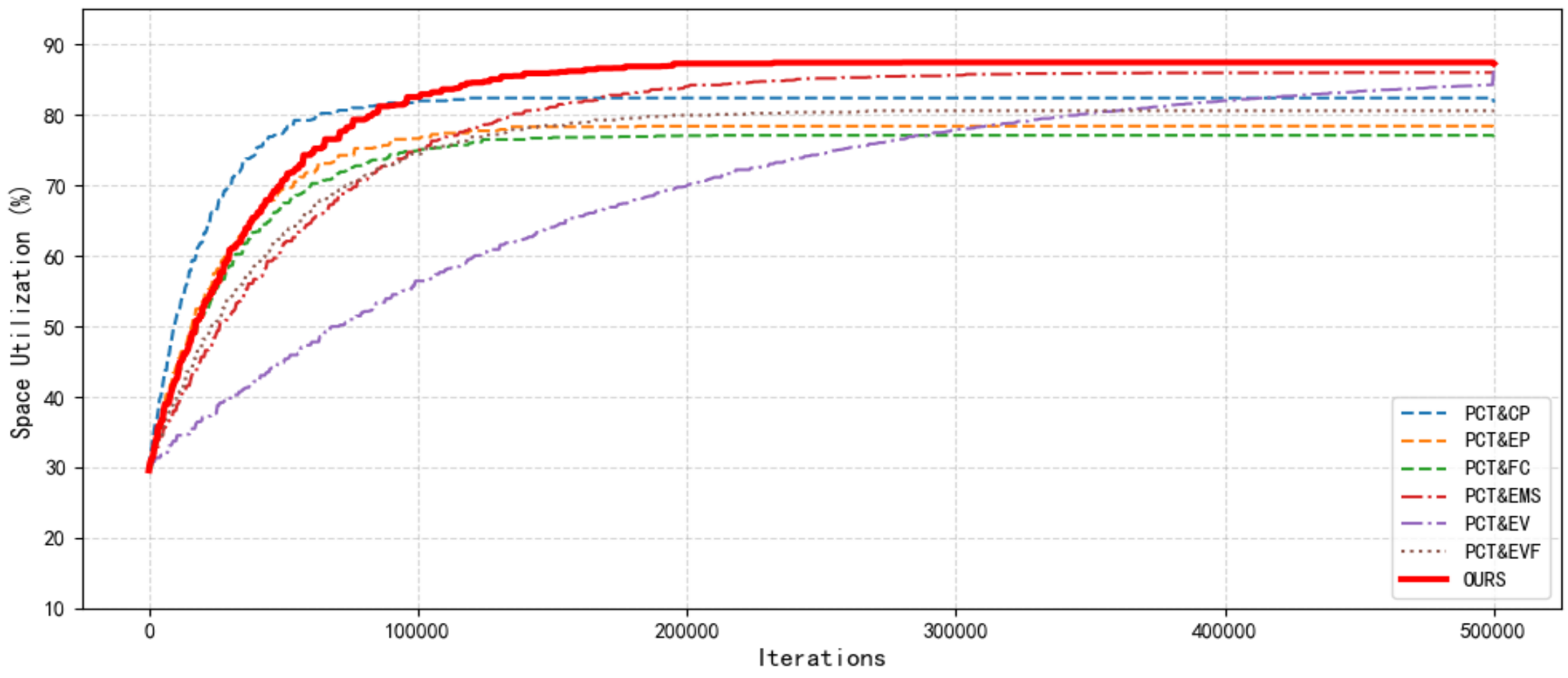

(b) Convergence curves of space utilization under Setting 2

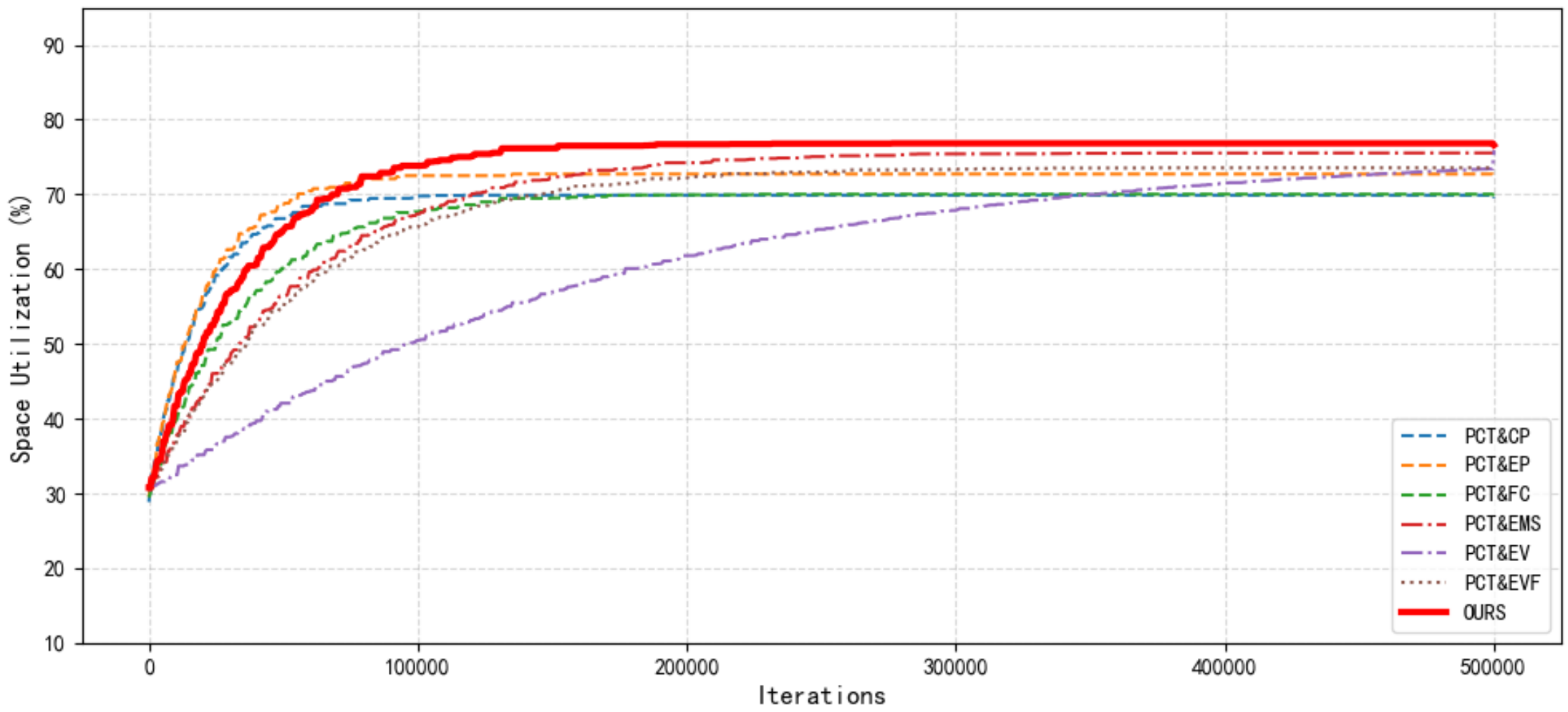

(c) Convergence curves of space utilization under Setting 3

Fig. 11 Comparison of Convergence Curves of Different Methods Under Various Datasets

To verify the effectiveness of the proposed method, comparative experiments are conducted with other algorithms under three different test settings: Setting1, Setting2 and Setting3.

The core evaluation metrics include packing utilization, variance and the number of packed items. The results are shown in Table 3. In terms of packing utilization, the proposed method significantly outperforms all comparison algorithms under the three settings and achieves the globally optimal performance. Specifically, the proposed method reaches 80.2% in Setting1, 87.3% in Setting2, and 76.6% in Setting3. Compared with the strong PCT-series baseline algorithms, the proposed method further improves utilization by 0.9%～4.3%, demonstrating the superiority of diffusion models in modeling high-dimensional complex packing action distributions and exploring better stacking layouts. Compared with traditional heuristic methods such as BR and LSAH, the utilization improvement of the proposed method exceeds 20%, fully showing the absolute advantage of reinforcement learning-based deep decision-making methods in spatial utilization efficiency. In terms of algorithm stability, by analyzing the variance (Var.) index, it can be seen that the proposed method maintains extremely low output fluctuations while achieving high utilization. The data shows that the variances of the proposed method are 4.2 in Setting1, 2.1 in Setting2, and 4.5 in Setting3, respectively. This index value is at the same level as that of high-performance methods such as PCT&EMS and PCT&EV, and much lower than that of early heuristic methods like BR and LSAH. This indicates that the proposed algorithm can not only achieve higher space utilization, but also generate packing results with better consistency and robustness, effectively solving the problems of large strategy fluctuations and unreliable output results in complex packing scenarios. In terms of packing efficiency index, the proposed method also performs prominently. In Setting1, Setting2, and Setting3, the number of packed items of the proposed method reaches 31.5, 34.2, and 30.3, respectively, all ranking first among all comparison methods. This result shows that in the limited container space, the proposed method can not only reduce gaps by optimizing the placement posture, but also efficiently accommodate a larger number of items, further verifying the practical value of the proposed framework in high-capacity and high-throughput packing scenarios.

Tab. 3 Comparison of Relevant Data of Different Methods

| | Setting1 | | | Setting2 | | | Setting3 | | |
|---|---|---|---|---|---|---|---|---|---|
| Method | Uti. | Var. | Num. | Uti. | Var. | Num. | Uti. | Var. | Num. |
| BR[20] | 51.2% | 7.40 | 20.90 | 56.7% | 6.60 | 22.60 | 48.9% | 10.70 | 19.5 |
| Ha et al.[36] | 52.1% | 20.1 | 20.6 | 59.9% | 10.4 | 23.8 | 51.9% | 20.2 | 20.6 |
| LSAH[37] | 53.2% | 6.90 | 21.10 | 65.0% | 6.10 | 25.60 | 52.4% | 12.20 | 20.7 |
| Wang &Hauser[38] | 57.6% | 11.5 | 24.1 | 66.1% | 8.4 | 25.9 | 56.5% | 11.2 | 22.3 |
| MACS[27] | 59.5% | 7.90 | 23.20 | 50.8% | 8.80 | 20.10 | 57.7% | 10.60 | 22.6 |
| DBL[39] | 60.5% | 8.8 | 23.8 | 70.6% | 7.9 | 27.8 | 60.5% | 8.9 | 23.8 |
| Zhao et al. | 72.0% | 5.30 | 27.90 | 70.3% | 4.30 | 27.40 | 59.6% | 5.40 | 23.1 |
| TOP[40] | 76.0% | 4.2 | 29.4 | 85.3% | 2.1 | 32.8 | 75.7 | 4.6 | 29.2 |
| MPC[41] | 57.7% | -- | 36.8 | | | | | | |
| GOPT[41] | 0.533 | -- | 34.2 | | | | | | |
| PCT&CP | 71.2% | 5.30 | 27.30 | 81.8% | 2.00 | 31.30 | 69.5% | 5.40 | 26.7 |
| PCT&EP | 72.5% | 5.60 | 28.20 | 78.1% | 3.80 | 30.30 | 72.2% | 5.80 | 27.9 |
| PCT&FC | 73.9% | 4.50 | 28.50 | 76.9% | 3.30 | 29.70 | 69.8% | 5.30 | 27.1 |
| PCT&EMS | 78.3% | 4.20 | 29.60 | 86.0% | 1.90 | 33.00 | 75.5% | 4.70 | 29.2 |
| PCT&EV | 79.1% | 4.2 | 30.2 | 87.1% | 3.1 | 32.8 | 75.9% | 4.6 | 29.1 |
| PCT&EVF | 75.7% | 4.8 | 29.2 | 80.5% | 2.9 | 31.0 | 73.5% | 4.6 | 28.4 |
| OURS | 80.2% | 4.2 | 31.5 | 87.3% | 2.1 | 34.2 | 76.6% | 4.5 | 30.3 |

In summary, the method proposed in this paper achieves transcendence over existing methods in three core dimensions: high utilization rate, high stability, and a large number of packed items. Whether facing the high-difficulty Setting 2 or the highly diverse Setting 3, the algorithm can output stable and optimal comprehensive performance. This fully indicates that the reinforcement learning framework combined with the diffusion model can better adapt to the 3D packing needs under different scenarios, providing an efficient and robust new solution for solving complex online packing problems.

To systematically verify the comprehensive performance of the method proposed in this paper under the continuous action space, this study conducts comparative experiments with multiple methods and scenarios, and the results are shown in Table 4. The method proposed in this paper achieves a balanced optimal performance in the three core indicators, topping the list in two scenarios and closely following the optimal in one scenario, making it the method with the strongest comprehensive competitiveness in this comparison. All well-performing DRL methods fully outperform traditional heuristic algorithms, verifying the core advantages of deep reinforcement learning in solving 3D bin packing problems under the continuous action space. In contrast, the poor performance of the GD method highlights the necessity of designing dedicated methods for bin packing problems. Compared with the currently optimal PCT series methods, the method proposed in this paper further upgrades the performance on their basis, and further improves the effect of bin packing optimization.

Figure 12 visualizes the 3D packing layouts of the proposed algorithm under three instances with 30, 32, and 34 items, respectively. As the number of items increases, the container space utilization rises steadily. The layouts are compact, regular, and stably stacked without structural imbalance, which intuitively verifies the high space utilization and strong robustness of the algorithm under different item scales, demonstrating its adaptability to dynamically changing packing requirements.

Tab. 4 Model Performance in Continuous Action Space

| Method | | Setting1 | | | Setting2 | | | Setting3 | | |
|---|---|---|---|---|---|---|---|---|---|---|
| | | Uti. | Var. | Num. | Uti. | Var. | Num. | Uti. | Var. | Num. |
| Heu. | BR[20] | 40.9% | 7.4 | 16.1 | 45.3% | 5.2 | 17.8 | 40.9% | 7.3 | 16.1 |
| | Ha et al.[36] | 43.9% | 14.2 | 17.2 | 46.1% | 6.8 | 18.1 | 43.9% | 14.2 | 17.2 |
| | LSAH[37] | 48.3% | 12.1 | 18.7 | 58.7% | 4.6 | 22.8 | 48.4% | 12.2 | 18.8 |
| DRL | GD | 5.6% | - | 2.2 | 7.5% | - | 2.9 | 5.2% | - | 2.1 |
| | PCT&EMS | 65.3% | 4.4 | 24.9 | 66.3% | 2.3 | 27.0 | 66.6% | 3.3 | 25.3 |

| | | | | | | | | | |
|---|---|---|---|---|---|---|---|---|---|
| PCT&EV | 65.4% | 3.3 | 25.0 | 65.0% | 2.6 | 26.4 | 65.8% | 3.6 | 25.1 |
| OURS | 66.5% | 3.9 | 24.4 | 67.4% | 3.6 | 27.0 | 64.5% | 3.8 | 26.4 |

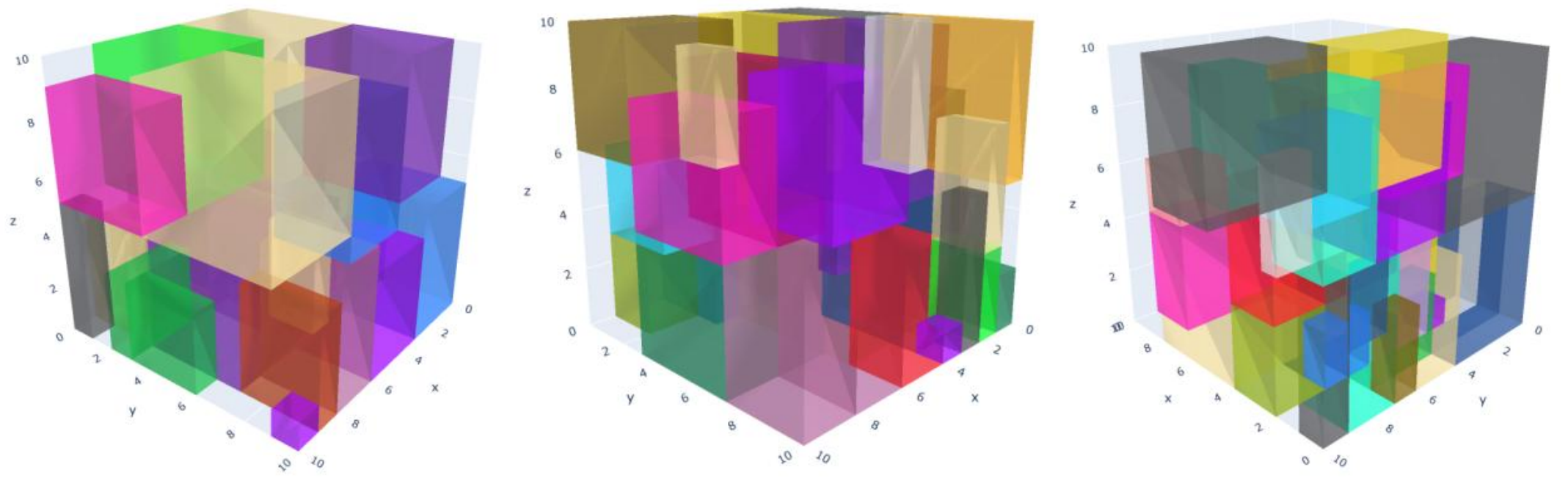

(a) Item=30 (b) Item=32 (c) Item=34

Fig. 12 Comparison of Visualized 3D Packing Layouts Under Different Instances

To comprehensively evaluate the trade-off between computational efficiency and packing performance among different online 3D bin packing algorithms, this paper systematically compares the runtime performance of various methods under two typical constraints: discrete solution space and continuous solution space. The relevant experimental results are shown in Table 5.

From the time consumption comparison, it can be seen that traditional heuristic algorithms such as BR and LSAH achieve the lowest running time under both types of solution spaces due to their fast rule-based decision logic. However, they suffer from obvious bottlenecks in packing space utilization and are difficult to adapt to high-demand industrial packing scenarios.

In contrast, deep reinforcement learning methods achieve a significant improvement in packing utilization at the cost of a controllable increase in computational overhead through deep modeling and iterative optimization of the high-dimensional action space, showing stronger layout optimization ability. Nevertheless, some baseline methods suffer from excessive time consumption and insufficient real-time performance, which makes it difficult to meet the requirements of engineering deployment.

Tab. 5 Performance Comparison of Different Algorithms on Online 3D-BPP under Various Settings

| Method | | Setting1 | | Setting2 | | Setting3 | |
|---|---|---|---|---|---|---|---|
| | | Disc. | Cont. | Disc. | Cont. | Disc. | Cont. |
| Heuristic | Random | $4.59\times10^{-2}$ | – | $2.03\times10^{-2}$ | – | $4.62\times10^{-2}$ | – |
| | Wang &Hauser | $4.76\times10^{-2}$ | – | $3.01\times10^{-2}$ | – | $4.55\times10^{-2}$ | – |
| | DBL | $5.58\times10^{-2}$ | – | $1.87\times10^{-2}$ | – | $5.44\times10^{-2}$ | – |
| | BR[20] | $1.50\times10^{-2}$ | $1.69\times10^{-2}$ | $1.74\times10^{-2}$ | $1.76\times10^{-2}$ | $1.42\times10^{-2}$ | $1.62\times10^{-2}$ |
| | Ha et al.[36] | $5.89\times10^{-3}$ | $1.48\times10^{-2}$ | $3.39\times10^{-3}$ | $7.17\times10^{-3}$ | $4.86\times10^{-3}$ | $1.38\times10^{-2}$ |
| | LSAH[37] | $1.22\times10^{-2}$ | $1.44\times10^{-2}$ | $4.98\times10^{-3}$ | $7.02\times10^{-3}$ | $1.14\times10^{-2}$ | $1.33\times10^{-2}$ |

| Method | | Setting1 | | Setting2 | | Setting3 | |
|---|---|---|---|---|---|---|---|
| | | Disc. | Cont. | Disc. | Cont. | Disc. | Cont. |
| DRL | MACS[27] | $2.68\times10^{-2}$ | – | $3.00\times10^{-2}$ | – | $2.79\times10^{-2}$ | – |
| | Zhao et al.[42] | $5.51\times10^{-2}$ | – | $1.33\times10^{-2}$ | – | $3.31\times10^{-2}$ | – |
| | PCT&CP | $8.43\times10^{-3}$ | $1.61\times10^{-2}$ | $7.36\times10^{-3}$ | $1.52\times10^{-2}$ | $8.79\times10^{-3}$ | $1.73\times10^{-2}$ |
| | PCT&EP | $1.22\times10^{-2}$ | $3.73\times10^{-2}$ | $1.13\times10^{-2}$ | $1.57\times10^{-2}$ | $1.25\times10^{-2}$ | 3.65×10−2 |
| | PCT&EMS | $1.77\times10^{-2}$ | $4.11\times10^{-2}$ | $9.49\times10^{-3}$ | $2.36\times10^{-2}$ | $1.80\times10^{-2}$ | 3.08×10−2 |
| | PCT&EV | $2.66\times10^{-2}$ | $4.46\times10^{-2}$ | $1.25\times10^{-2}$ | $3.21\times10^{-2}$ | $2.61\times10^{-2}$ | 4.38×10−2 |
| | OURS | $1.55\times10^{-2}$ | $4.7\times10^{-2}$ | $1.03\times10^{-2}$ | $1.88\times10^{-2}$ | $2.04\times10^{-2}$ | 3.34×10−2 |

The method proposed in this paper achieves a better balance between performance and efficiency by integrating the global optimization capability of diffusion reinforcement learning and the fast decision-making advantages of heuristic strategies:

Compared with traditional heuristic algorithms, the proposed method achieves a significant improvement of more than 20% in packing utilization at the cost of minimal additional time overhead. While ensuring the real-time performance of decision-making, it completely breaks through the performance upper limit of heuristic methods.

Compared with pure DRL baseline methods, by introducing heuristic prior constraints, the proposed method effectively compresses the invalid exploration space. It not only maintains a competitive space utilization rate but also significantly reduces the running time, realizing efficient reasoning under both discrete and continuous solution spaces.

The above results fully demonstrate that the method proposed in this paper not only has the high optimization performance of deep reinforcement learning but also retains the advantage of low time consumption of heuristic algorithms. It meets the real-time requirements while ensuring the packing effect, and is more suitable for the engineering application needs of online 3D bin packing in actual industrial scenarios.

## 5.2 Ablation studies

To verify the actual gain of the heuristic module on model performance under different action spaces, this paper designs ablation comparison experiments under both discrete and continuous action spaces. The results are shown in Table 6.

Tab. 6 Ablation Results of Heuristic Module

| Experimental Scenario | Method | Setting1 | | | Setting2 | | | Setting3 | | |
|---|---|---|---|---|---|---|---|---|---|---|
| | | Uti. | Var. | Num. | Uti. | Var. | Num. | Uti. | Var. | Num. |
| Continuous | without heuristic method (random sampling) | 62.8% | 6.7 | 23.2 | 63.5% | 4.1 | 25.1 | 61.9% | 5.8 | 23.7 |
| | with heuristic method | 66.5% | 3.9 | 24.4 | 67.4% | 3.6 | 27.0 | 64.5% | 3.8 | 26.4 |
| Discrete | without heuristic | 75.5% | 6.8 | 28.1 | 82.2% | 4.5 | 30.7 | 71.8% | 7.2 | 27.5 |
| | method (random sampling) | | | | | | | | | |
| | with heuristic method | 80.2% | 4.2 | 31.5 | 87.3% | 2.1 | 34.2 | 76.6% | 4.5 | 30.3 |

In the continuous action space, after introducing the heuristic strategy, the model's space utilization rate increased by an average of 3.3% under the three experimental settings, the result variance decreased by an average of 37%, and the number of packed items increased by an average of 1.2. In the discrete action space, the performance gain is more significant: the space utilization rate increased by an average of 4.9%, the result variance decreased by an average of 45%, and the number of packed items increased by an average of 3.1. The experimental results fully demonstrate that the heuristic candidate generation strategy can effectively compress the action search space, avoid invalid exploration, and form a complement with the spatial feature extraction capability of the graph network. It can significantly improve the packing performance and stability under different action spaces, verifying the necessity and core value of this module in the 3D bin packing model.

To explore the optimization effect of the height penalty mechanism on packing performance, this paper conducts ablation experiments under both discrete and continuous action spaces, and the results are shown in Table 7. In the continuous space, after adding the height difference constraint, the average space utilization rate of the model increases by 4.2%, the average variance decreases by 48%, and the average number of packed items increases by 1.8. In the discrete space, the improvement is equally significant: the average utilization rate increases by 4.2%, the variance decreases by 51%, and the number of packed items increases by 2.5. The experimental results show that the height difference constraint can effectively avoid stacking imbalance and loose layout, and significantly improve the packing stability and space utilization rate under different action spaces, which proves the necessity and important role of this module.

To explore the effect of modeling the spatial topological relationship between items on improving the performance of 3D bin packing, and to verify the core value of the graph network module in the overall framework, this paper designs ablation comparison experiments under both discrete and continuous action spaces. By comparing the experimental results before and after introducing the graph network, the optimization effect of this module on layout perception and feature extraction is quantitatively analyzed, and the results are shown in Table 8.

Tab. 7 Ablation Experimental Results of the Height Penalty Mechanism

| Experimental Scenario | Method | Setting1 | | | Setting2 | | | Setting3 | | |
|---|---|---|---|---|---|---|---|---|---|---|
| | | Uti. | Var. | Num. | Uti. | Var. | Num. | Uti. | Var. | Num. |
| Continuous | no height difference constraint (only basic constraint) | 62.1% | 7.5 | 23.0 | 63.8% | 6.2 | 25.5 | 60.7% | 8.1 | 24.2 |
| | With height difference constraint | 66.5% | 3.9 | 24.4 | 67.4% | 3.6 | 27.0 | 64.5% | 3.8 | 26.4 |
| Discrete | no height difference constraint (only basic | 76.0% | 8.5 | 29.0 | 83.1% | 4.5 | 31.5 | 72.5% | 9.0 | 28.1 |
| | constraint) | | | | | | | | | |
| | With height difference constraint | 80.2% | 4.2 | 31.5 | 87.3% | 2.1 | 34.2 | 76.6% | 4.5 | 30.3 |

Tab. 8 Ablation Results of Graph Network Module

| Experimental Scenario | Method | Setting1 | | | Setting2 | | | Setting3 | | |
|---|---|---|---|---|---|---|---|---|---|---|
| | | Uti. | Var. | Num. | Uti. | Var. | Num. | Uti. | Var. | Num. |
| Continuous | Without GCN | 63.2% | 5.8 | 23.5 | 64.1% | 4.7 | 25.9 | 61.3% | 6.5 | 24.8 |
| | With GCN | 66.5% | 3.9 | 24.4 | 67.4% | 3.6 | 27.0 | 64.5% | 3.8 | 26.4 |
| Discrete | Without GCN | 76.8% | 7.8 | 29.2 | 83.5% | 4.3 | 31.6 | 72.7% | 8.2 | 27.8 |
| | With GCN | 80.2% | 4.2 | 31.5 | 87.3% | 2.1 | 34.2 | 76.6% | 4.5 | 30.3 |

In the continuous action space, after introducing the graph network, the average space utilization rate of the model under the three experimental settings is increased by 3.3%, the average result variance is reduced by 33%, and the average number of packed items is increased by 1.2. In the discrete action space, the performance gain is more significant: the average space utilization rate is increased by 3.7%, the average result variance is reduced by 47%, and the average number of packed items is increased by 2.5.

The experimental results fully show that the graph network module can effectively encode the spatial correlation and topological structure between items, significantly improve the model's global perception ability of complex 3D layouts, and provide high-quality feature support for upper-level heuristic candidate generation and height difference constraints. This module can achieve dual improvements in packing performance and stability under different action spaces, further verifying its necessity and core contribution in the 3D bin packing model.

## 6 Conclusions

This paper proposes a diffusion reinforcement learning method integrated with spatial heuristic strategies to improve the packing performance of diffusion reinforcement learning. On the basis of diffusion reinforcement learning, this paper introduces classical bin packing heuristic rules as prior knowledge to perform feature fusion with the actions generated by the diffusion model; at the same time, a multi-dimensional spatial information representation module is constructed to enhance the model's ability to understand the layout structure; a hybrid reward function is designed to guide strategy optimization. The diffusion reinforcement learning algorithm integrated with spatial heuristic information has significantly improved layout rationality, space utilization and robustness in complex scenarios. However, the algorithm mainly packs cuboid goods and is difficult to generalize to the packing scenario of irregular objects. Future research will combine visual technology to realize the modeling and packing processing of special-shaped goods, so as to improve the scenario adaptability of the algorithm.

# Acknowledgement

This work was supported by the Natural Science Foundation of Shandong Province (ZR2024MF036), Teaching Reform Project of Shandong University (Y2025034). We thank Hang Zhao for releasing their open-source code of deep reinforcement learning baselines at https://github.com/alexfrom0815/Online-3D-BPP-DRL, which helped us to implement the comparison experiments efficiently.

## Refereces